\documentclass[lettersize,journal]{IEEEtran}
\usepackage{amsmath,amsfonts}
\usepackage{algorithmic}
\usepackage{algorithm}
\usepackage{array}
\usepackage[caption=false,font=normalsize,labelfont=sf,textfont=sf]{subfig}
\usepackage{textcomp}
\usepackage{stfloats}
\usepackage{url}
\usepackage{verbatim}
\usepackage{graphicx}
\usepackage{cite}
\usepackage[numbers,sort&compress]{natbib}

\usepackage{tabularx}
\usepackage{booktabs}
\usepackage{enumitem}
\usepackage{subfig}
\usepackage{multirow}
\usepackage{adjustbox}
\usepackage{threeparttable}
\usepackage{xcolor}

\begin{document}




\title{From Camera-Based Sensing to Reasoning: A Comprehensive Review Toward Proactive Vulnerable Road User Safety}




\author{%
    Shucheng Zhang\textsuperscript{1}, 
    Yan Shi\textsuperscript{1}, 
    Bingzhang Wang\textsuperscript{1},
    Yuang Zhang\textsuperscript{1}, 
    Muhammad Monjurul Karim\textsuperscript{1}, 
    Kehua Chen\textsuperscript{1},
    Chenxi Liu\textsuperscript{2},
    Mehrdad Nasri\textsuperscript{1}, 
    Yinhai Wang\textsuperscript{1,*}%
    \thanks{\textsuperscript{1}Department of Civil and Environmental Engineering, University of Washington, Seattle, WA 98105, USA.}%
    \thanks{\textsuperscript{2}Department of Civil and Environmental Engineering, University of Utah,  Salt Lake City, UT 84112, USA.}%
    \thanks{*Corresponding author. E-mail: yinhai@uw.edu}%
}

\markboth{Journal of \LaTeX\ Class Files,~Vol.~14, No.~8, August~2021}%
{Shell \MakeLowercase{\textit{et al.}}: A Sample Article Using IEEEtran.cls for IEEE Journals}


\maketitle

\begin{abstract}
Ensuring the safety of vulnerable road users (VRUs), such as pedestrians and cyclists, remains a critical challenge, as conventional infrastructure-based measures are often insufficient in dynamic urban environments. Recent advances in learning-based visual sensing systems have created new opportunities for more adaptive and context-aware VRU protection. This paper presents a comprehensive review of camera-based methods for proactive VRU safety, with a focus on developments over the past five years. Unlike prior surveys that primarily emphasize detection, we organize the literature into three interconnected components: visual perception (detection and classification), motion modeling (tracking and trajectory prediction), and behavior understanding (intent recognition and reasoning). These components form a unified hierarchical pipeline that enables early risk anticipation and timely intervention. Furthermore, this survey systematically incorporates emerging AI paradigms, including Vision Transformers (ViTs), Large Language Models (LLMs), and diffusion models, highlighting their roles in representation learning, uncertainty modeling, and semantic reasoning. Finally, we identify four key challenges specific to VRU safety, including data scarcity, behavioral uncertainty, edge deployment efficiency, and real-world sensing constraints, and discuss corresponding research directions. This work provides a unified foundation for the development of reliable, scalable, and deployable VRU safety systems.
\end{abstract}


\section{Introduction}
\IEEEPARstart{R}{oad} transportation safety is a global concern that requires sustained attention. Compared to vehicle occupants, other road users, such as pedestrians and cyclists, are more vulnerable to injury in traffic incidents \citep{khan2024advancing}, and are therefore referred to as VRUs. According to the \citet{who2023road}, over 1.19 million people die each year from road traffic crashes globally, with more than half of those fatalities being VRUs. In the United States, the situation is similarly concerning. In 2021, VRUs accounted for nearly 20\% of the 42,915 traffic fatalities, increasing 13\% over 2020 \citep{usdotsafety2022}.

\begin{figure}[h]
  \centering
  \subfloat[\scriptsize Number of Publications: AI for VRU Safety]{%
    \includegraphics[width=0.3\textwidth]{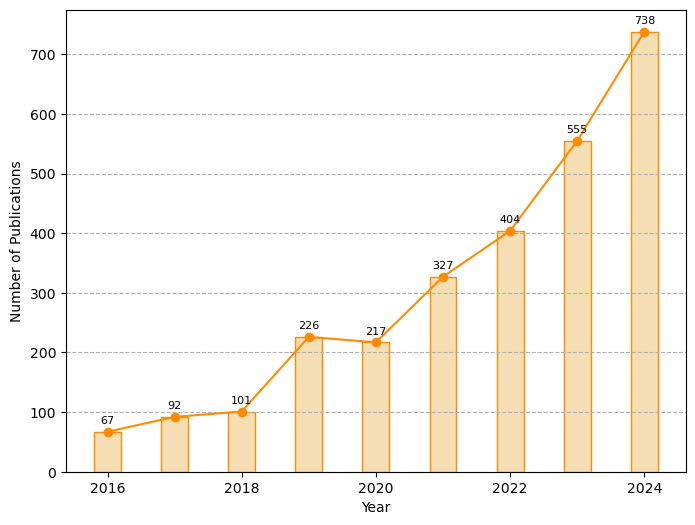}%
    \label{fig:trenda}
  }
  \hspace{0.05\textwidth} 
  \subfloat[\scriptsize Number of Publications: AI for Pedestrian Safety]{%
    \includegraphics[width=0.3\textwidth]{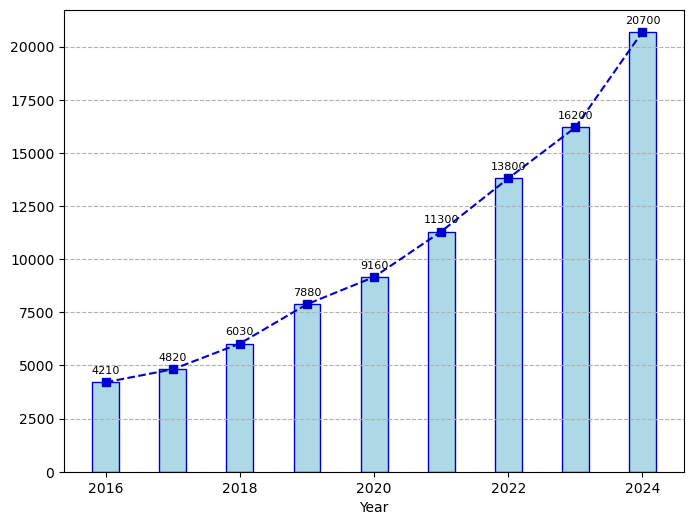}%
    \label{fig:trendb}
  }
  \caption{Publication Trends in AI Research for VRU and Pedestrian Safety (2016–2024) Based on Google Scholar Search Using the Keywords “AI for VRU Safety” and “AI for Pedestrian Safety”}
  \label{fig:trend}
\end{figure}

Despite decades of effort, passive safety measures, such as signage, speed enforcement, and pedestrian-prioritized infrastructure design, have not been sufficient to effectively protect VRUs. These approaches lack the flexibility and situational awareness required to prevent accidents in complex and rapidly changing environments \citep{hafner2013cooperative}. In recent years, advances in learning-based and multimodal sensing techniques have enabled AI-driven sensing systems that offer a more proactive, scalable, and cost-effective approach to enhancing VRU safety, addressing the limitations of traditional methods \citep{zhang2025vulnerable}. These systems can support high-accuracy detection, behavior modeling, and intent understanding, enabling both vehicles and infrastructure to better anticipate and respond to potential hazards before they escalate into collisions. Figure \ref{fig:trend} illustrates the publication trends related to learning-based methods in VRU safety and pedestrian safety from 2016 to 2024. As shown in the figure, research on learning-based methods has grown rapidly in recent years, reflecting the increasing adoption of data-driven techniques in this domain. In parallel, the research focus has gradually expanded from pedestrian-specific studies to the broader concept of VRUs, encompassing a wider range of non-motorized traffic participants. This shift highlights the growing need for more comprehensive and inclusive safety solutions.

\captionsetup[subfloat]{font=footnotesize}

\begin{table*}[ht]
\begin{scriptsize}
\centering
\caption{Summary of Existing Review Papers on learning-based VRU Sensing}
\begin{tabularx}{\textwidth}{@{}p{4.2cm} p{0.9cm} p{2.3cm} X X@{}}
\toprule
\textbf{Paper Name} & \textbf{Year} & \textbf{Publication} & \textbf{Highlight} & \textbf{Limitations} \\
\midrule

Vulnerable Road User Detection for Roadside-Assisted Safety Protection: A Comprehensive Survey \citep{zhang2025vulnerable} & 2025 & Applied Sciences & Focuses on infrastructure-based VRU detection, combining traditional and recent approaches. & Primarily addresses detection; lacks discussion on advanced models and multi-task perception. \\
\midrule
Roadside Sensor Systems for Vulnerable Road User Protection: A Review of Methods and Applications \citep{zhang2025roadside} & 2025 & IEEE Access & Summarizes methods and applications of roadside sensors in VRU protection. & Offers limited analysis from the algorithmic and model development perspective. \\
\midrule
Vulnerable Road User Detection and Safety Enhancement: A Comprehensive Survey \citep{silva2024vulnerable} & 2024 & arXiv & Surveys VRU safety technologies, including sensors, communication systems, and data fusion. & Provides limited discussion of visual perception tasks beyond detection. \\
\midrule
Advancing Vulnerable Road Users Safety: Interdisciplinary Review on V2X Communication and Trajectory Prediction \citep{abdi2024advancing} & 2024 & IEEE TITS & Reviews V2X technologies and trajectory prediction models for enhancing VRU safety. & Focuses primarily on communication and prediction; lacks discussion on visual perception and multimodal AI models. \\
\midrule
Occlusion Handling and Multi-Scale Pedestrian Detection Based on Deep Learning: A Review \citep{li2022occlusion} & 2022 & IEEE Access & Highlights deep learning-based approaches for handling occlusion and scale variation in pedestrian detection. & Focuses solely on detection under occlusion; lacks coverage of modern models and broader visual tasks. \\
\midrule
From Handcrafted to Deep Features for Pedestrian Detection: A Survey \citep{cao2021handcrafted} & 2021 & IEEE TPAMI & Provides a comprehensive view of pedestrian detection, from handcrafted features to deep CNN-based methods. & Does not cover state-of-the-art perception models or other visual understanding tasks. \\

\bottomrule
\end{tabularx}
\label{tab:review_summary}
\end{scriptsize}
\end{table*}

Several review papers have explored the role of deep learning in improving VRU safety, as summarized in Table \ref{tab:review_summary}. However, most of these works concentrate primarily on detection tasks and their associated methodologies, offering limited consideration of other vision-based tasks that are equally important for comprehensive VRU protection. Key areas such as trajectory and intent prediction, which can provide richer context for interpreting VRU behaviors, are often overlooked. Moreover, many of these reviews focus primarily on traditional CNN-based or attention-based models, recent breakthroughs in ViTs, LLMs, and Diffusion Models have introduced novel capabilities in multimodal input processing and high-level behavior understanding. Despite their promise, these cutting-edge approaches remain largely unexamined in the current literature. These limitations highlight the need for a more comprehensive and up-to-date review that captures the full scope of modern AI advancements in VRU proactive protection.

To address the limitations in existing literature, this paper presents a comprehensive review of recent advances in vision-based tasks for VRU safety enhancement, with a particular focus on developments from the past five years.
The specific contributions of this survey are as follows:

\begin{enumerate}
\item \textbf{Unified multi-stage perspective:} We provide a comprehensive review of camera-based methods for VRU safety, organized into a unified pipeline consisting of visual perception, motion modeling, and behavior understanding. This structured view moves beyond prior surveys that focus on isolated tasks and highlights the progression from fundamental perception to practical deployment.

\item \textbf{Camera-centric analysis:} This survey explicitly centers on camera-based visual sensing from both vehicular and infrastructure perspectives, the most widely deployed modality in real-world systems, enabling a more coherent and in-depth analysis compared to multi-sensor reviews.

\item \textbf{Integration of emerging AI paradigms:} We systematically examine recent advances in modern AI architectures and discuss their roles in representation learning, uncertainty modeling, and reasoning for VRU safety.

\item \textbf{Proactive and deployment-oriented insights:} We identify four key challenges, VRU data scarcity, behavioral uncertainty, edge efficiency, and real-world sensing constraints, and discuss future directions toward reliable, scalable, and proactive VRU safety systems.
\end{enumerate}

The remainder of this paper is organized as follows. In Section \ref{sec:background}, we introduce the basic concepts related to the definition and categorization of VRUs, as well as an overview of safety-relevant visual tasks. In Section \ref{sec:core_tasks}, we review three core vision-based tasks for VRU safety and summarize recent advancements in each area. Section \ref{sec:challenges} identifies four major open challenges related to data, model, and hardware limitations, and discusses emerging strategies to address them. Finally, Section \ref{sec:conclusion} concludes the paper.

\section{Foundational Concepts}
\label{sec:background}

\subsection{Definition and Subclasses of VRU }

VRUs are individuals who face a high risk of injury or death in traffic environments due to their lack of physical protection and exposure to fast-moving vehicles. While the term has traditionally referred to pedestrians and cyclists, its modern scope has expanded to include a wider range of non-motorized or lightly protected road users \citep{nsc2025vulnerable}. Despite this broadening, a clear and standardized definition of each VRU subclass remains lacking, often leading to inconsistencies in research and safety assessments \citep{zhang2025vulnerable}. In response to this gap, the 2024 Intersection Safety Challenge, hosted by \citet{usdotsafetychallenge}, introduced a comprehensive categorization and formal definitions for various VRU classes and subclasses. Based on these guidelines, Table \ref{tab:vru_classification} summarizes the classification scheme along with the visual sensing challenges of each group. Overall, VRUs are generally small in size, exhibit highly dynamic behaviors, and have inconsistent speed profiles, making them inherently challenging to detect and track \citep{zhang2025vulnerable}. 

Beyond these general characteristics, the diversity among VRU types introduces additional sensing challenges due to variations in appearance, movement patterns, and contextual interactions with the environment. For example, pedestrians display a wide range of poses, clothing styles, and levels of occlusion, especially in crowded urban settings. Cyclists and scooter users travel at higher speeds and follow more flexible trajectories, requiring perception models that can accommodate rapid motion and irregular paths. Users of mobility aids such as canes or walkers often exhibit similar silhouettes when viewed from surveillance cameras, making them difficult to distinguish, particularly for infrastructure-mounted sensors with limited vertical field of view. Moreover, non-motorized device users carrying strollers, luggage, or umbrellas are often misclassified due to their visual similarity to standard pedestrians. These additional objects are hard to distinguish and can obscure body outlines, further reducing recognition accuracy \citep{damen2011detecting}.

\begin{table*}[!ht]
\begin{footnotesize}
\caption{Detailed Classification of VRUs and Associated Sensing Challenges}
\label{tab:vru_classification}
\begin{center}
\begin{tabularx}{0.8\textwidth}{@{}p{2.5cm} p{5cm} p{8cm}@{}}
\toprule
\textbf{Class} & \textbf{Subclasses} & \textbf{Primary Sensing Challenges} \\
\midrule

Pedestrian 
  & \begin{itemize}[nosep,leftmargin=*]
      \item Child
      \item Adult
      \item Elder
    \end{itemize}
  & \begin{itemize}[nosep,leftmargin=*]
      \item High variability in appearance and motion
      \item Frequent occlusion in urban scenes
      \item Different movement and reaction patterns
    \end{itemize} \\

\midrule

Cyclist  
  & \begin{itemize}[nosep,leftmargin=*]
      \item Manual Bicycle
      \item Motorized Bicycle
    \end{itemize}
  & \begin{itemize}[nosep,leftmargin=*]
      \item Fast and often non-linear motion
      \item Occlusion by frames or helmets
      \item Visual confusion with motorcycles and scooters
    \end{itemize} \\

\midrule

Non-Motorized Device 
  & \begin{itemize}[nosep,leftmargin=*]
      \item Cane, Crutches, Walker
      \item Stroller, Umbrella
      \item Cardboard Box, Luggage
    \end{itemize}
  & \begin{itemize}[nosep,leftmargin=*]
      \item Unusual shapes and attachments deform body contours
      \item Occlusion of core body region
      \item Often treated as background or misclassified
    \end{itemize} \\

\midrule

Wheelchair 
  & \begin{itemize}[nosep,leftmargin=*]
      \item Manual Wheelchair
      \item Motorized Wheelchair
    \end{itemize}
  & \begin{itemize}[nosep,leftmargin=*]
      \item Low visual profile leading to detection failure
      \item Ambiguity with small vehicles or static obstacles
    \end{itemize} \\

\midrule

Scooter and Skateboard
  & \begin{itemize}[nosep,leftmargin=*]
      \item Manual Scooter
      \item Motorized Scooter
      \item Skateboard
    \end{itemize}
  & \begin{itemize}[nosep,leftmargin=*]
      \item Dynamic standing postures and rapid speed variation
      \item Challenging to track over time due to agile motion
      \item Difficult to distinguish from pedestrians in motion
    \end{itemize} \\

\bottomrule
\end{tabularx}
\end{center}

\end{footnotesize}
\end{table*}

\subsection{Vision-Based Tasks for VRU Safety}
Camera-based sensing systems have become a cornerstone of modern transportation safety due to their affordability, ease of deployment, and ability to capture rich semantic information \citep{silva2024vulnerable}. Compared to other sensing modalities such as lidar and radar, cameras provide detailed visual context necessary for fine-grained tasks like pose estimation. Additionally, camera infrastructure is already widely deployed across urban environments and integrated into vehicles, making it a practical and scalable backbone for VRU safety systems.

The configuration and placement of cameras play a critical role in determining their sensing capabilities. Vehicle-mounted cameras offer an ego-centric perspective aligned with the driver’s viewpoint, but are limited by restricted fields of view and motion dynamics. In contrast, infrastructure-mounted cameras provide a global perspective, enabling long-range detection, crosswalk monitoring, and multi-agent tracking. These complementary viewpoints are essential for comprehensive VRU sensing, and this review considers both to provide a comprehensive understanding of camera-based systems.

\begin{figure*}[!ht]
	\centering
	\includegraphics[width=.95\textwidth]{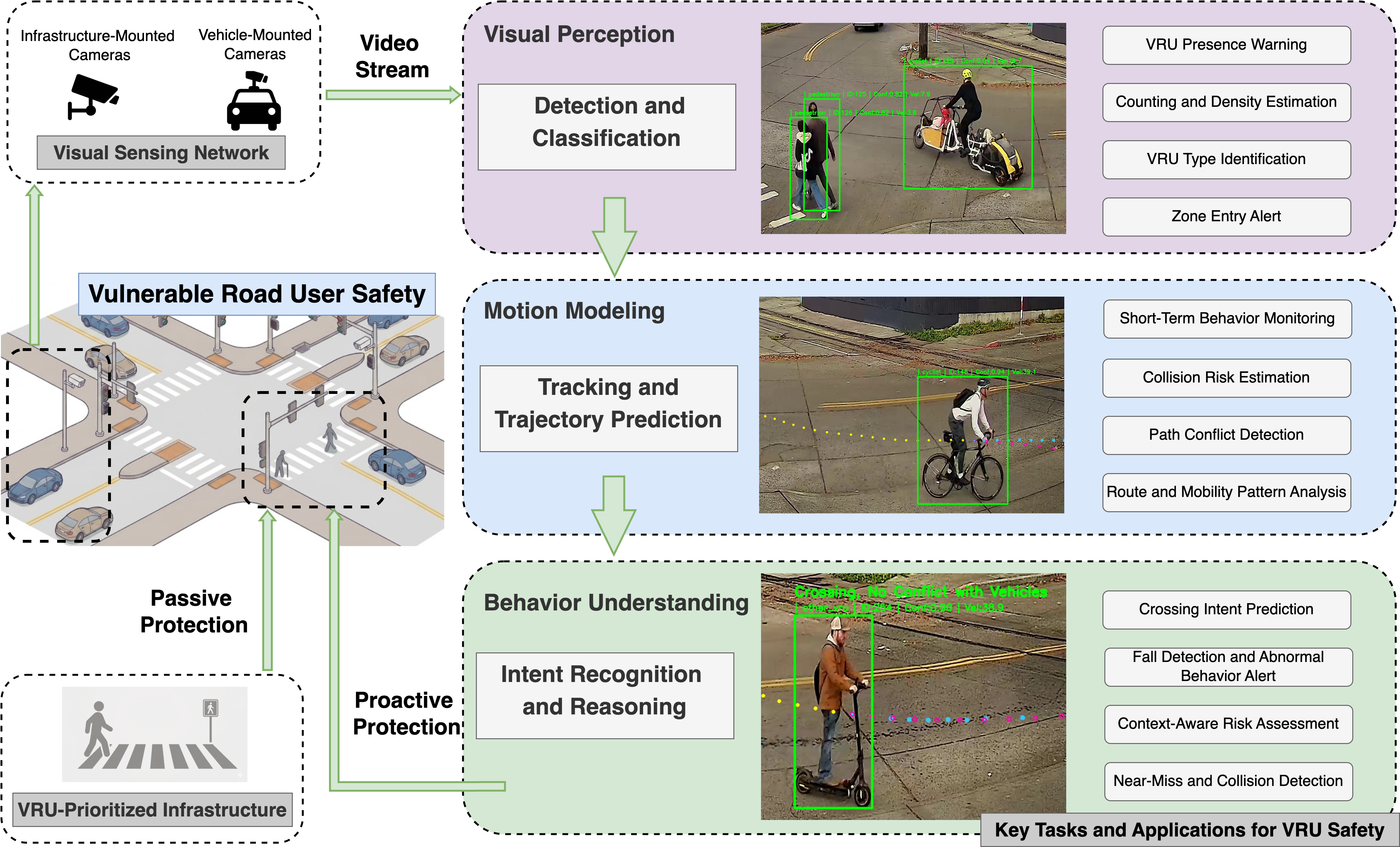}
	\caption{Framework of vision-based tasks for proactive VRU safety. Video streams from infrastructure or vehicle cameras are processed through a hierarchical pipeline consisting of visual perception, motion modeling, and behavior understanding. The example visualizations illustrate intermediate outputs across stages: bounding boxes indicate VRU detection results, points represent VRU past and predicted trajectories, and text annotations indicate inferred intent and reasoning outcomes. These progressively richer representations enable downstream applications, ultimately supporting proactive safety interventions.}
	\label{fig:safety structure}
\end{figure*}

Building on these advantages, recent advances in computer vision have expanded the role of camera-based systems beyond traditional object detection toward more comprehensive and proactive safety applications. While early research primarily focused on detection as the core perception capability, such a narrow perspective overlooks additional components required for context-aware and predictive safety systems. To address this gap, we organize existing vision-based methods into a unified pipeline consisting of three interconnected components: visual perception, motion modeling, and behavior understanding, as illustrated in Figure~\ref{fig:safety structure}.

Within this pipeline, visual perception (detection and classification) serves as the foundation by localizing and identifying VRUs from video streams. Motion modeling (tracking and trajectory prediction) captures temporal dynamics, enabling continuous monitoring and short-term forecasting of VRU movements. Building on these representations, behavior understanding (intent recognition and reasoning) interprets higher-level actions by incorporating contextual information such as interactions with vehicles, road geometry, and traffic signals.

These components provide progressively richer representations of VRUs, evolving from appearance cues to motion patterns and ultimately to semantic behavior interpretation. Their integration enables traditionally fragmented vision tasks to operate within a coherent framework, supporting end-to-end reasoning from perception to decision-making. As a result, intelligent transportation systems can better anticipate potential risks and understand complex interactions, forming the foundation for proactive safety interventions. By combining passive infrastructure with active, camera-based intelligence, modern VRU protection systems can achieve more adaptive and context-aware safety beyond the limitations of conventional approaches.

\section{Core Vision-Based Methodologies for VRU Safety}
\label{sec:core_tasks}

\subsection{Visual Perception: Detection and Classification}

The evolution of image-based VRU detection has broadly followed the trajectory of general object detection, progressing from traditional hand-crafted feature pipelines \citep{Dalal, Viola2004Robust} to modern learning-based representations. Despite these advances, VRU detection remains particularly challenging due to factors such as dense crowds, severe occlusions, and low-light environments. These challenges necessitate specialized techniques beyond standard detection frameworks and continue to drive active research in this area. In this section, we review recent advances in detection models and their adaptation to VRU scenarios, followed by specialized approaches designed to address domain-specific challenges. Table~\ref{tab:vru_detection_methods} summarizes representative methods developed over the past five years.

\subsubsection{General Visual VRU Detection Methods}
Monocular RGB cameras, as one of the most widely deployed sensing modalities in both vehicles and infrastructure, provide rich semantic information for real-time VRU detection. The development of detection models has evolved from convolutional architectures toward transformer-based, multimodal, and generative paradigms, with continuous improvements in accuracy, efficiency, and robustness under real-world conditions. Early and CNN-based models remain widely adopted due to their computational efficiency and suitability for real-time deployment. For instance, EfficientDet \citep{tan2020efficientdetscalableefficientobject} introduced compound scaling and a BiFPN architecture to balance model complexity and accuracy, while Sparse R-CNN \citep{sun2021sparse} leveraged learnable proposals and sparse attention to improve detection in cluttered scenes. PP-YOLOE \citep{xu2022pp} further optimized real-time detection performance for VRUs, and ConvNeXtV2 \citep{woo2023convnext} enhanced generalization through modernized convolutional design. InternImage \citep{wang2023internimage} adopted deformable convolutions to improve localization of small or partially occluded objects. 

Building upon these advances, ViT and transformer-based detection models have become increasingly prominent due to their ability to capture global context and long-range dependencies. Deformable DETR \citep{zhu2020deformable} improves detection of small and densely distributed objects through deformable attention mechanisms, while SwinV2 \citep{liu2022swin} introduces hierarchical feature representations for dense pedestrian environments. Recent YOLO variants, including YOLOv11 \citep{khanam2024yolov11} and YOLOv12 \citep{tian2025yolov12}, integrate transformer components and advanced attention mechanisms to enhance feature aggregation and reduce latency. RT-DETR \citep{zhao2024detrs} further bridges the gap between accuracy and efficiency by enabling real-time end-to-end detection, while hybrid architectures such as FasterViT \citep{hatamizadeh2023fastervit} combine convolutional and transformer operations for improved performance. Lightweight transformer variants, such as EdgeViT \citep{chen2022edgevit} and MaxViT \citep{tu2022maxvit}, also facilitate deployment on edge devices by balancing computational efficiency and representational power.

More recently, large-scale multimodal models, particularly VLMs, have introduced a new paradigm for VRU detection by incorporating high-level semantic reasoning. For instance, VLPD \citep{liu2023vlpdcontextawarepedestriandetection} leveraged vision-language semantic self-supervision to enhance contextual understanding in crowded scenes, while MS-VLMDet \citep{dai2025ms} introduced multi-scale cross-modal feature alignment to improve detection across varying pedestrian sizes. MSCoTDet \citep{kim2024mscotdet} further incorporated language-driven chain-of-thought reasoning to guide multi-modal fusion, demonstrating improved robustness under adverse conditions. These approaches highlight the growing trend of integrating high-level semantic reasoning into detection pipelines.

Diffusion models represent another emerging paradigm that formulates detection as a generative process. DiffusionDet \citep{chen2023diffusiondet} approached detection as a denoising task, showing robustness to noisy or incomplete annotations. Extending this direction, diffusion-based methods have also been applied to data generation and feature enhancement for pedestrian detection. For example, Diffusion Dataset Generation \citep{farley2023diffusion} utilized generative diffusion models to synthesize realistic pedestrian samples and reduce the sim-to-real gap, while anisotropic diffusion-based frameworks \citep{barbu2025novel} improved detection performance through multi-scale feature enhancement and noise suppression. These developments suggest that diffusion mechanisms can contribute both at the model level and the data level to improve VRU detection in complex environments.

While recent models have significantly improved detection for pedestrians and cyclists, other VRU types, such as e-scooter riders, skateboarders, and individuals using mobility aids, remain underexplored. Only a few studies have started addressing this gap. \citet{yang2023improved} proposed a modified version of YOLOv5 with task-specific improvements for VRU detection, resulting in better recall and precision in surveillance footage. \citet{Gilroy_2022} developed a benchmark and detection framework specifically for e-scooter riders, tackling challenges like occlusion and visual similarity with other VRUs. \citet{shourov2021deep} used deep learning to analyze interactions between skateboarders and pedestrians, emphasizing the need for fine-grained classification. \citet{apurv2021detection} focused on detecting e-scooter riders in natural scenes, designing a visual pipeline that could distinguish them from similar-looking users under occlusion. Finally, a system proposed by \citet{vasquez2017deep} for mobility aids detection combined visual and depth information to recognize individuals using canes, walkers, or wheelchairs. 

A key barrier to broader research on VRUs is the lack of labeled datasets covering these less common categories. To address this, few-shot and open-set detection approaches are emerging as effective solutions. VLMs, in particular, allow for training-free extensions by using text prompts to describe unseen categories. Grounding DINO \citep{liu2024grounding} aligned visual and textual representations to enable open-set detection, successfully identifying rare VRUs like wheelchair users. YOLO-World \citep{cheng2024yolo} integrated language-conditioned learning into the YOLO framework, enabling flexible detection of new VRU types without retraining. These approaches offer a scalable and adaptive path forward for inclusive VRU safety systems.

\subsubsection{Detection in Dense and Occluded Environments}

Detecting VRUs in crowded and occluded scenes remains a major challenge for visual perception systems. In real-world environments, pedestrians are often partially blocked or closely packed, leading to detection failures by standard models that rely on full-body visibility and rigid object representations. To address these limitations, recent methods have focused on visible part reasoning, feature recovery, and crowd-aware modeling.

A common strategy is to emphasize visible body regions. The Mask-Guided Attention Network (MGAN) \citep{pang2019maskguidedattentionnetworkoccluded} uses coarse segmentation masks to focus attention on the visible portions of a pedestrian. Similarly, Representative Region NMS (R2NMS) \citep{huang2020nms} enhances detection under occlusion by learning robust representations from visible parts and refining them with specialized non-max suppression. On the other hand, some methods aim to recover occluded features. FeatComp++ employs adversarial training to hallucinate the missing parts of a partially visible pedestrian, aligning its features with those of fully visible instances \citep{zhang2024imagine}. MAPD \citep{wang2021mapd} introduces discriminative feature learning for multi-attribute pedestrian detection, which improves robustness when individuals are partially visible or occluded by others. 

Additionally, crowd-aware detectors also explore alternative object representations and anchor strategies. Beta R-CNN proposed by \citet{xu2020beta} replaces bounding boxes with probabilistic beta distribution masks and uses a soft suppression strategy, improving overlap resolution in dense groups. One Proposal, Multiple Predictions (OPMP) \citep{chu2020detectioncrowdedscenesproposal} introduces multiple predictions per anchor to detect overlapping individuals and improves matching with Earth Mover’s Distance loss. PedHunter \citep{chi2020pedhunter} is an occlusion-robust detector tailored for crowded environments and tested on large pedestrian datasets. APPM \citep{liu2021adaptive} introduces pattern-parameter matching to improve the detection of pedestrians with diverse occlusion patterns. OAF-Net \citep{li2022oaf} removes anchor boxes and applies occlusion-aware focal loss, enabling better learning under dense occlusions. AutoPedestrian  \citep{tang2021autopedestrian} combines convolutional backbones with neural architecture search to optimize augmentation and loss functions for occlusion-heavy datasets. HeadHunters \citep{sundararaman2021tracking} focuses on detecting and tracking pedestrian heads, which are more frequently visible in dense crowds. Optimal Proposal Learning (OPL) \citep{song2023optimal} further improves crowd detection by optimizing proposal quality in an end-to-end training framework. Lastly, transformer-based approaches have also been adapted for crowd detection. DETR for Crowd Pedestrian \citep{lin2021detrcrowdpedestriandetection} applies DETR to dense scenes by redesigning the transformer head to better handle occlusion and high-density pedestrian layouts.

Given that such crowded environments are prevalent in urban transportation systems, these advancements are critical for developing reliable and inclusive AI-powered VRU safety solutions.

\begin{table*}[!ht]
\begin{footnotesize}
\begin{center}
\caption{Representative Methods for VRU Detection and Classification (P: Pedestrian; C: Cyclist; O: Other VRU Types)}
\begin{tabularx}{0.90\textwidth}{l l l l l l}
\toprule
\textbf{Method} & \textbf{Backbone} & \textbf{Year} & \textbf{Target} & \textbf{Dataset} & \textbf{Publication} \\
\midrule
\multicolumn{6}{l}{\textbf{General Object Detection Methods}} \\
EfficientDet~\citep{tan2020efficientdetscalableefficientobject}        & CNN                             & 2020 & P, C            & COCO~\citep{lin2014microsoftcoco}                      & CVPR \\
Sparse R-CNN~\citep{sun2021sparse}        & Hybrid             & 2021 & P, C            & COCO                      & CVPR \\
PP-YOLOE~\citep{xu2022pp}             & CNN                             & 2022 & P, C            & COCO                      & ArXiv \\
ConvNeXtV2~\citep{woo2023convnext}             & CNN                             & 2023 & P, C            & COCO                      & CVPR \\
InternImage~\citep{wang2023internimage}         & CNN                             & 2023 & P, C            & COCO                      & CVPR \\
Deformable DETR~\citep{zhu2020deformable}     & Transformer                            & 2020 & P, C            & COCO                      & ArXiv \\
SwinV2~\citep{liu2022swin}              & Transformer                            & 2022 & P, C            & COCO                      & CVPR \\
YOLOv11~\citep{khanam2024yolov11}             & Hybrid                     & 2024 & P, C            & COCO                      & ArXiv \\
YOLOv12~\citep{tian2025yolov12}             & Hybrid         & 2025 & P, C            & COCO                      & ArXiv \\
RT-DETR~\citep{zhao2024detrs}             & Transformer                            & 2024 & P, C            & COCO                      & CVPR \\
FasterViT~\citep{hatamizadeh2023fastervit}           & Hybrid                     & 2023 & P, C            & COCO, ImageNet~\citep{ImageNet}            & ICLR \\
EdgeViT~\citep{chen2022edgevit}             & Transformer                            & 2022 & P, C            & ImageNet, COCO            & ECCV \\
MaxViT~\citep{tu2022maxvit}              & Transformer                            & 2022 & P, C            & ImageNet, COCO            & ECCV \\
VLPD ~\citep{liu2023vlpdcontextawarepedestriandetection}& VLM          & 2023 & P, C & CityPersons~\citep{zhang2017citypersons}              & CVPR \\
MS-VLMDet ~\citep{dai2025ms} & VLM          & 2025 & P & CityPersons              & ITSC  \\
MSCoTDet ~\citep{kim2024mscotdet} & VLM          & 2025 & P  & FLIR~\citep{flir2025adas}, CVC-14~\citep{gonzalez2016pedestrian}              &TCSVT \\
DiffusionDet~\citep{chen2023diffusiondet}        & Diffusion                        & 2023 & P, C            & COCO, CrowdHuman~\citep{shao2018crowdhuman}          & CVPR \\

\midrule
\multicolumn{6}{l}{\textbf{Detection Methods in Low-Light Conditions}} \\

Cyclic Fuse-and-Refine~\citep{zhang2020multispectral}          & CNN      & 2020 & P                    & KAIST~\citep{choi2018kaist}, FLIR                    & ICIP \\
MBNet~\citep{zhou2020improving}                           & CNN      & 2020 & P                    & KAIST, CVC-14                  & ECCV \\
GAFF~\citep{zhang2021guided}                            & CNN      & 2021 & P                    & KAIST, FLIR                    & WACV \\
CFT~\citep{qingyun2021cross}                             & Hybrid      & 2021 & P                    & FLIR, LLVIP~\citep{jia2021llvip}                    & ArXiv \\
IT-MN~\citep{zhuang2021illumination}                           & CNN      & 2021 & P                    & KAIST                          & ArXiv \\
UFF+UCG~\citep{kim2021uncertainty}                         & CNN      & 2022 & P                    & KAIST, CVC-14                  & TCSVT \\
ASG-LPF~\citep{li2022pedestrian}                             & CNN      & 2022 & P                    & KAIST                          & Optik \\
MS-DETR~\citep{xing2024ms}                         & Transformer     & 2023 & P                    & KAIST, CVC-14, LLVIP           & TITS \\
MFDs-YOLO + i-IAN~\citep{hsia2023all}               & CNN      & 2023 & P, C                 & KAIST, FLIR                    & Electronics \\
CMM~\citep{kim2024causal}                             & CNN      & 2024 & P                    & KAIST, FLIR, CVC-14      & CVPR \\
\midrule
\multicolumn{6}{l}{\textbf{Detection Methods in Dense and Occluded Environments}} \\
MGAN~\citep{pang2019maskguidedattentionnetworkoccluded}                            & Hybrid      & 2019 & P                    & CityPersons, Caltech~\citep{dollar2009pedestrian}           & ArXiv \\
PedHunter~\citep{chi2020pedhunter}                       & CNN      & 2020 & P                    & CityPersons, Caltech, COCO  & AAAI \\
DETR for Crowd Pedestrian~\citep{lin2021detrcrowdpedestriandetection}       & Transformer     & 2020 & P                    & CrowdHuman, CityPersons        & ArXiv \\
R2NMS~\citep{huang2020nms}                           & CNN      & 2020 & P                    & CrowdHuman, CityPersons        & CVPR \\
OPMP~\citep{chu2020detectioncrowdedscenesproposal}                            & CNN      & 2020 & P                    & CrowdHuman, CityPersons, COCO  & CVPR \\
Beta R-CNN~\citep{xu2020beta}                      & CNN      & 2020 & P                    & CrowdHuman, CityPersons        & NeurIPS \\
APPM~\citep{liu2021adaptive}                            & CNN      & 2021 & P                    & Caltech, CityPersons           & AAAI \\
MAPD~\citep{wang2021mapd}                            & CNN      & 2021 & P                    & CrowdHuman, CityPersons        & Neurocomputing \\
HeadHunters~\citep{sundararaman2021tracking}                     & CNN      & 2021 & P                    & CroHD~\cite{sundararaman2021tracking}                           & CVPR \\
AutoPedestrian\citep{tang2021autopedestrian}                  & CNN & 2021 & P                    & CrowdHuman, CityPersons        & TIP \\
OAF-Net~\citep{li2022oaf}                         & CNN      & 2022 & P                    & CrowdHuman, Caltech, CityPersons & TITS \\
OPL~\citep{song2023optimal}       & CNN      & 2023 & P                    & CrowdHuman, TJU-Ped~\citep{Pang_2021}, Caltech   & CVPR \\
FeatComp++~\citep{zhang2024imagine}                      & CNN      & 2024 & P                    & CityPersons, Caltech, CrowdHuman & ArXiv \\

\midrule
\multicolumn{6}{l}{\textbf{Detection Methods for Other Types of VRUs}} \\
Detection of Mobility Aids~\citep{vasquez2017deep}     & CNN      & 2017 & O           & Custom dataset                 & ArXiv \\
Skateboarder–Pedestrian Detection~\citep{shourov2021deep}& CNN    & 2021 & O           & Custom dataset                 & MDPI \\
Detection of E-scooter Riders~\citep{apurv2021detection}   & CNN      & 2021 & O          & Custom dataset                 & ArXiv \\
E-Scooter Rider Detection~\citep{Gilroy_2022}      & CNN      & 2022 & O           & Custom dataset                 & ArXiv \\
YOLOv5 for VRU Detection~\citep{yang2023improved}        & CNN      & 2023 & O          & Custom dataset                 & Sensors \\
Grounding DINO~\citep{liu2024grounding}      & VLM          & 2024 & P, C, O & LVIS~\citep{gupta2019lvisdatasetlargevocabulary}, COCO              & ECCV \\
YOLO-World~\citep{cheng2024yolo}         & VLM           & 2024 & P, C, O & LVIS, COCO               & CVPR \\
\bottomrule
\end{tabularx}
\label{tab:vru_detection_methods}
\end{center}
\end{footnotesize}
\end{table*}

\subsubsection{Detection in Low-Light Conditions}
Ensuring the safety of VRUs under low-light or nighttime conditions is a critical yet challenging task for camera-based perception systems. Traditional RGB cameras often struggle in the dark due to reduced contrast and increased noise. To overcome these limitations, researchers have developed a variety of deep learning approaches that enhance detection by either fusing complementary sensing modalities or adapting model behavior based on illumination \citep{ghari2024pedestrian}.

Among these solutions, multispectral fusion, particularly the combination of RGB and thermal infrared imagery, has become the most widely adopted strategy. Transformer-based architectures such as MS-DETR \citep{xing2024ms} employ dual CNN backbones for RGB and thermal inputs, along with modality-specific transformer encoders and a multimodal decoder. This model incorporates modality-balanced optimization to ensure equal contribution from both input streams. Similarly, the Cross-modality Fusion Transformer (CFT) \citep{qingyun2021cross} aligns region features from each modality to reduce misalignment and improve detection robustness. Several CNN-based fusion models have also achieved strong performance in low-light scenarios. Guided Attentive Feature Fusion (GAFF) \citep{zhang2021guided} uses attention modules within a two-stream CNN to highlight the strengths of each modality. Cyclic Fuse-and-Refine \citep{zhang2020multispectral} introduces cyclic fusion blocks designed to address spatial shifts between RGB and thermal images. MBNet \citep{zhou2020improving} handles modality imbalance using dual-modality attention fusion and illumination-aware feature alignment, improving detection consistency across lighting conditions.

Beyond fusion strategies, some models incorporate illumination awareness directly into the network architecture. \citet{zhuang2021illumination} proposed the Illumination and Temperature-aware Multispectral Network (IT-MN), which uses a Fusion Weight Network to adjust the contributions of RGB and infrared features based on ambient light and temperature. This adaptive behavior allowed the model to rely more on thermal imagery in dark conditions, leading to more effective detection. Similarly, Double-Stream Multispectral Network (DSMN) \citep{hsia2023all} integrates two YOLO-based sub-networks and an illumination-aware module to estimate scene brightness and dynamically fuse the two modalities. This approach proved more effective in handling fluctuating lighting from sources such as vehicle headlights or street lamps compared to static fusion methods. Further developments by \citet{li2019illumination} and \citet{li2022pedestrian} focused on controlling information flow based on lighting conditions. The former proposed an Illumination-Aware Faster R-CNN that uses a learned brightness estimator to switch between visible and thermal detection branches. The latter expanded this idea with the Adaptive Soft-Gated Light Perception Fusion (ASG-LPF) module, which continuously adjusts the fusion process based on the output of a light perception sub-network. This allowed the model to transition smoothly between daytime, dusk, and nighttime conditions, thereby improving stability and detection accuracy across varied lighting environments.

\subsection{Motion Modeling: Tracking and Trajectory Prediction}

With the increasing adoption of intelligent systems in transportation, modeling the motion of VRUs has become a critical research problem. Motion modeling typically involves two components: tracking, which estimates the temporal evolution of VRU states from observations, and trajectory prediction, which forecasts future movements based on historical dynamics. Together, these components provide the foundation for proactive safety applications such as near-miss detection, collision avoidance, and behavior analysis.

\subsubsection{Tracking}
Tracking VRUs across video frames from a single camera is essential for understanding their movement patterns and ensuring proactive safety in transportation systems. However, it presents unique challenges such as occlusion, identity switches, and limited field-of-view, especially in dynamic or crowded environments. 

The foundation of modern single-camera multi-object tracking (MOT) lies in the tracking-by-detection paradigm. Classical methods such as SORT and DeepSORT remain widely used in this field. SORT \citep{bewley2016simple} proposed a light-weight pipeline that combines a Kalman filter for motion prediction and the Hungarian algorithm for data association, offering real-time performance suitable for on-board systems. DeepSORT \citep{wojke2017simple} extended the framework by integrating a deep appearance descriptor trained for re-identification. This enhancement significantly improved identity consistency over longer sequences and under partial occlusions, making DeepSORT one of the most widely adopted baselines in both academic and industrial applications for pedestrian tracking.

In recent years, more unified and robust learning-based approaches have been proposed in the field. Notably, CenterTrack \citep{zhou2020tracking} introduced a joint detection and tracking framework that predicts object centers and their offsets from previous frames, allowing end-to-end training without explicit association stages. Similarly, FairMOT \citep{zhang2021fairmot} proposed a fully convolutional anchor-free network that performs object detection and identity embedding simultaneously, achieving a balance between detection accuracy and ID preservation. Other methods like JDE \citep{wang2020towards}, QDTrack \citep{pang2021quasi}, and CSTrack \citep{liang2022rethinkingcompetitiondetectionreid} adopt a one-shot strategy to simultaneously learn object localization and appearance features, thus avoiding hand-crafted association logic. More recently, ByteTrack \citep{zhang2021bytetrack} achieved a breakthrough by proposing that even low-confidence detections can improve tracking robustness, especially in dense or occluded scenarios. Innovations in motion modeling have also emerged: OC-SORT \citep{cao2023observation} refines Kalman-based prediction by incorporating virtual trajectory updates during occlusion, while BoT-SORT \citep{aharoni2022bot} combines strong appearance cues with camera motion compensation. Additionally, transformer-based architectures such as TrackFormer \citep{chen2021trackformer}, TransTrack \citep{sun2020transtrack}, and MOTR \citep{zeng2022motr} have been proposed to model long-term spatial-temporal dependencies, eliminating explicit association steps and achieving state-of-the-art results on MOT benchmarks.

In summary, the field of single-camera VRU tracking has evolved from modular, rule-based pipelines to highly integrated deep learning frameworks that unify detection, re-identification, and temporal modeling. These methods have significantly improved tracking accuracy and identity preservation, particularly under challenging conditions. By leveraging both spatial and appearance cues, modern tracking systems now enable real-time, high-reliability applications that are critical for VRU protection in intelligent transportation systems. 

\subsubsection{Trajectory Prediction}
Accurately modeling human motion, particularly for trajectory prediction, is inherently more challenging than for other road users such as vehicles \citep{golchoubian2023pedestrian}. These challenges arise from three main factors. First, human movement is less constrained by rigid dynamical rules, allowing pedestrians to change direction and speed with high flexibility. Second, predicting pedestrian behavior is more complex due to the absence of strict lane constraints and the strong influence of contextual factors such as surrounding traffic and infrastructure. Third, errors in motion modeling can have severe safety implications, as VRUs are more vulnerable to injury in collision scenarios. These challenges highlight the importance of capturing both spatial-temporal interactions and behavioral cues for reliable prediction. Early approaches primarily relied on physics-based models and classical machine learning techniques, which often assumed simplified motion patterns and handcrafted features \citep{9756903}. In contrast, recent advances have been driven by learning-based methods that automatically capture complex spatial and temporal dependencies from data, leading to substantial improvements in prediction accuracy and robustness.

Recurrent architectures have been widely adopted for modeling temporal dependencies in VRU trajectory prediction, particularly Long Short-Term Memory (LSTM) and Gated Recurrent Unit (GRU) networks. These models effectively capture sequential motion patterns in pedestrian trajectories by encoding temporal dynamics over time. For example, \citet{li2020recurrent} introduced a recurrent attention mechanism that dynamically reweights different time steps to emphasize behaviorally salient motion patterns and mitigate speed variations. To further enhance temporal representation, \citet{zhang2022ai} integrated convolutional GRUs with graph-based contextual encoding, enabling more effective propagation of temporal features in dynamic traffic scenes. Similarly, \citet{tang2023trajectory} proposed a multi-scale temporal modeling strategy by incorporating dilated temporal convolutions prior to LSTM-based generation, thereby extending the temporal receptive field without increasing model complexity. In addition, \citet{mo2022multi} introduced type-aware RNN encoders to capture heterogeneous motion dynamics across different VRU categories. Collectively, these RNN-based approaches provide adaptive and context-aware solutions for modeling sequential VRU motion, particularly in scenarios with strong temporal dependencies.

To explicitly model interactions among multiple agents, GNNs have emerged as a fundamental framework for VRU trajectory prediction. GNN-based methods represent traffic participants as nodes and their interactions as edges, enabling structured modeling of spatial relationships. For instance, GSTCN \citep{sheng2022graph} combines graph convolution for spatial interaction modeling with convolutional layers for temporal feature extraction. Building upon this, FJMP \citep{rowe2023fjmp} constructs sparse directed acyclic graphs to model future interaction dynamics, enabling scalable and consistent joint trajectory prediction. To capture higher-order social behaviors, GroupNet \citep{xu2022groupnet} introduces multiscale hypergraph representations that extend beyond pairwise interactions to model group-level dynamics.

Building upon these foundations, ViT-based architectures have become increasingly dominant due to their ability to model global spatial-temporal dependencies through self-attention. Unlike RNNs, which process sequences sequentially, or GNNs, which rely on predefined graph structures, transformer-based models provide a unified framework for jointly modeling interactions across time, agents, and scene context. For example, the Scene Transformer \citep{ngiam2021scene} adopts a scene-centric representation and applies attention across temporal, agent, and environmental dimensions to generate consistent multi-agent trajectory predictions. The MTR framework \citep{shi2022mtr} leverages paired motion queries within a transformer decoder to simultaneously model goal-oriented and trajectory-specific behaviors, achieving state-of-the-art performance in multimodal prediction. In contrast, Obstacle-Transformer \citep{zhang2023obstacle} reduces reliance on explicit scene representations by modeling interactions directly from structured trajectory inputs, demonstrating the flexibility of attention-based reasoning. Additionally, iNATran \citep{chen2022vehicle} incorporates multi-attention mechanisms, including social, temporal, and cross-attention, within a non-autoregressive transformer framework, enabling efficient generation of multiple plausible future trajectories. Furthermore, several recent approaches integrate GNNs with transformer-based architectures, such as HiVT \citep{zhou2022hivt} and AI-TP \citep{zhang2022ai}, where graph-based encoders are used to model spatial relationships before applying attention mechanisms for temporal reasoning. This hybrid design combines the structured relational modeling capability of GNNs with the global reasoning power of transformers, further improving performance in complex multi-agent scenarios. 

Finally, generative models have rapidly emerged as a primary approach for modeling uncertainty and multimodality in VRU trajectory prediction. \citet{gu2022stochastic} first introduced Motion Indeterminacy Diffusion (MID), which explicitly models the transition from ambiguous walkable regions to determinate trajectories using a transformer-based diffusion process. Subsequent works have focused on improving the efficiency and practicality of diffusion models; for instance, \citet{mao2023leapfrog} proposed the Leapfrog Diffusion Model (LED), which employs a trainable initializer to skip multiple denoising steps and enables real-time stochastic prediction without sacrificing diversity. Beyond efficiency, recent studies have addressed the challenge of limited or incomplete observations. \citet{chen2023equidiff} introduced EquiDiff, a conditional diffusion framework with an SO(2)-equivariant transformer backbone to preserve geometric consistency while modeling social interactions via RNN and GNN encoders. To further handle extreme observation sparsity, \citet{li2023bcdiff} proposed a bidirectional consistent diffusion framework that jointly reconstructs missing historical trajectories and predicts future motion, while \citet{luo2025diffusion} extended this idea with uncertainty-aware dual diffusion processes that adapt noise based on reconstructed trajectory confidence.

In parallel, efforts have been made to enhance generalization and controllability. \citet{bae2024singulartrajectory} proposed a unified diffusion-based framework capable of handling diverse trajectory prediction settings, including stochastic, deterministic, and few-shot scenarios within a shared representation space. \citet{jiang2023motiondiffuser} further introduced a controllable multi-agent diffusion framework that enables the incorporation of external constraints during inference, facilitating more realistic and policy-compliant motion generation. Complementary to diffusion-based approaches, \citet{luo2024granp} leveraged a Neural Process framework to model uncertainty with spatial-temporal graph representations, while \citet{chen2024human} incorporated human-like decision-making through a diffusion-based planner coupled with inverse reinforcement learning. More recently, \citet{chib2025lg} explored the integration of large language models to generate high-level motion cues, highlighting the potential of combining generative models with semantic reasoning for trajectory prediction.

Table~\ref{tab:trajectory_benchmark_ethucy} presents a quantitative comparison of representative methods on the ETH-UCY pedestrian trajectory benchmark~\citep{eth2009,ucy2007}. Methods are grouped by evaluation protocol, as differences in observation window length and prediction strategy preclude direct cross-group comparison. Under the standard stochastic protocol (Group I), diffusion-based models consistently outperform RNN-based and GNN-based approaches, with average ADE decreasing from 0.49\,m~\citep{li2020recurrent} to 0.21\,m~\citep{bae2024singulartrajectory}, demonstrating the effectiveness of diffusion-based generation for multi-modal pedestrian trajectory prediction. These advances demonstrate how generative models are evolving from purely stochastic predictors to unified, controllable, and context-aware frameworks, offering more expressive and reliable representations of VRU motion under uncertainty.

\begin{table*}[!ht]
\renewcommand{\arraystretch}{1.3}
\begin{footnotesize}
\begin{center}
\caption{Quantitative comparison of representative methods on the ETH-UCY benchmark (ADE/FDE in meters, lower is better). All methods report minADE/minFDE with best-of-20 sampling. \textbf{Bold} indicates the best result per column within each group. Methods across groups use different observation protocols and are \textbf{not directly comparable}.}
\label{tab:trajectory_benchmark_ethucy}
\begin{tabularx}{0.97\textwidth}{@{}l c X X X X X X@{}}
\toprule
\textbf{Method} & \textbf{Obs.} & \textbf{ETH} & \textbf{Hotel} & \textbf{Univ} & \textbf{Zara1} & \textbf{Zara2} & \textbf{AVG} \\
\midrule
\multicolumn{8}{l}{\textit{Group I: Standard Stochastic Prediction (8-frame observation $\rightarrow$ 12-frame prediction)$^\dagger$}} \\
\midrule
Li et al.~\citep{li2020recurrent}
  & 8 & 0.79/1.60 & 0.44/0.91 & 0.57/1.26 & 0.34/0.72 & 0.29/0.62 & 0.49/1.02 \\
Xu et al. (GroupNet)~\citep{xu2022groupnet}
  & 8 & 0.46/0.73 & 0.15/0.25 & 0.26/0.49 & 0.21/0.39 & 0.17/0.33 & 0.25/0.44 \\
Gu et al. (MID)~\citep{gu2022stochastic}
  & 8 & 0.39/0.66 & 0.13/0.22 & 0.22/0.45 & 0.17/0.30 & 0.13/0.27 & 0.21/0.38 \\
Mao et al. (LED)~\citep{mao2023leapfrog}
  & 8 & 0.39/0.58 & \textbf{0.11/0.17} & 0.26/0.43 & 0.18/0.26 & \textbf{0.13/0.22} & 0.21/0.33 \\
Bae et al. (SingularTrajectory)~\citep{bae2024singulartrajectory}
  & 8 & \textbf{0.35/0.42} & 0.13/0.19 & \textbf{0.25/0.44} & \textbf{0.19/0.32} & 0.15/0.25 & \textbf{0.21/0.32} \\
\midrule
\multicolumn{8}{l}{\textit{Group II: Deterministic Prediction (8-frame observation $\rightarrow$ 12-frame prediction, single sample)$^\ddagger$}} \\
\midrule
Zhang et al. (Obstacle-TF)~\citep{zhang2023obstacle}
  & 8 & 0.65/1.39 & 0.27/2.08 & 0.53/1.43 & 0.39/0.79 & 0.28/0.67 & 0.42/1.27 \\
\midrule
\multicolumn{8}{l}{\textit{Group III: Instantaneous/Momentary Prediction (2-frame observation $\rightarrow$ 12-frame prediction)$^\S$}} \\
\midrule
Li et al. (BCDiff)~\citep{li2023bcdiff}
  & 2 & 0.30/0.56 & 0.13/0.20 & 0.25/0.52 & 0.18/0.37 & 0.14/0.31 & 0.19/0.39 \\
Luo et al. (Diffusion$^2$)~\citep{luo2025diffusion}
  & 2 & \textbf{0.29/0.45} & \textbf{0.10/0.13} & \textbf{0.26/0.47} & \textbf{0.17/0.37} & \textbf{0.14/0.26} & \textbf{0.19/0.33} \\
\bottomrule
\end{tabularx}

\vspace{4pt}
\raggedright
\footnotesize{
$^\dagger$ Standard leave-one-out cross-validation protocol~\citep{guta2018};
  observe 3.2\,s (8 frames at 2.5\,Hz), predict 4.8\,s (12 frames);
  minADE/minFDE computed over 20 sampled trajectories. \\
$^\ddagger$ Single deterministic prediction (no sampling);
  same 8-frame observation and 12-frame prediction horizon as Group~I,
  but outputs only one trajectory, making direct comparison with stochastic methods inappropriate. \\
$^\S$ Instantaneous/momentary observation setting: only 2 frames (0.8\,s) are observed before prediction;
  designed for scenarios where agents emerge suddenly (e.g., from occlusions);
  minADE/minFDE still computed over 20 samples,
  but lower errors relative to Group~I reflect the shorter observation window rather than superior model capacity.
}
\end{center}
\end{footnotesize}
\end{table*}

\subsection{Behavior Understanding: Intent Recognition and Reasoning}

Intent recognition and reasoning aim to provide a high-level understanding of VRU behavior by inferring future actions and the underlying decision-making processes. Compared to trajectory prediction, which focuses on estimating future motion, behavior understanding seeks to interpret why a VRU is likely to act in a certain way within a given context. These two aspects are closely related and are often modeled jointly or hierarchically in modern frameworks \citep{zhang2023pedestrian}. Importantly, behavior understanding plays a critical role in enabling proactive VRU safety, as it allows intelligent systems to anticipate potentially hazardous actions, such as sudden road crossing or inattentive movement, before they occur, thereby supporting early intervention and risk mitigation. Table~\ref{tab:vru_intent_methods} summarizes representative methods and their main contributions for intent recognition and reasoning across different model paradigms.

However, intent recognition and reasoning is inherently challenging due to the high variability of VRU behavior and its dependence on both internal factors (e.g., walking speed, attention level) and external conditions (e.g., traffic flow, signal phase, and road geometry). To address these challenges, existing approaches leverage diverse temporal or spacial information beyond trajectory data, including human pose sequences, traffic signals, vehicle motion, and other visual contents within the scene. This task is commonly formulated as a classification problem, where models predict discrete behavioral states~\citep{goldhammer2019intentions}. Recently, reasoning-driven methods have emerged as a promising direction for more robust and interpretable behavior understanding.

\subsubsection{Temporal and Interaction-Based Intent Recognition}
With the rise of deep learning, RNN-based approaches quickly became dominant in modeling VRU intent, leveraging their ability to capture temporal dependencies in sequential data. VRUNet~\citep{ranga2020vrunet} introduced a multi-task LSTM architecture that jointly predicts pedestrian actions, crossing intent, and future trajectories using a combination of 2D human pose sequences and semantic scene features. Building on this idea,~\citet{yao2021coupling} proposed a Coupled Intent-Action model that jointly predicts pedestrian actions and crossing intent, where future action predictions serve as priors to refine current intent estimation. Beyond pedestrian intent prediction, research on other VRUs is emerging. For example, \citet{bridgeman2023gesture} developed a system to recognize cyclist hand signals using pose-based features.

To further enhance temporal modeling, later works incorporated attention mechanisms. PIP-Net~\citep{azarmi2024pip} integrates recurrent and temporal attention modules to dynamically weigh kinematic and visual cues, improving robustness under occlusion and complex scenarios. Similarly, PCPA~\citep{kotseruba2021benchmark} introduced a hybrid attention-based framework that applies temporal attention within each modality stream and modality attention for multi-source fusion, improving both accuracy and interpretability. Extending multimodal fusion, MCIP~\citep{ham2022mcip} and CIPF~\citep{ham2023cipf} employ GRU-based encoders combined with attention mechanisms to integrate pose, trajectory, and vehicle-related features, achieving improved robustness under missing or noisy inputs. Despite these advances, RNN-based methods remain limited in explicitly modeling interactions among multiple agents and capturing long-range dependencies.

Graph-based architectures have been introduced to explicitly capture social and spatial interactions influencing VRU behavior. Early work such as Pedestrian Graph~\citep{cadena2019pedestrian} represents human pose as a graph, where joints are treated as nodes and body structures as edges, enabling intent prediction from skeletal dynamics. Pedestrian Graph+~\citep{cadena2022pedestrian} further improves computational efficiency, facilitating real-time deployment. Beyond single-agent modeling,~\citet{liu2020spatiotemporal} proposed a spatiotemporal GCN framework that constructs pedestrian-centric scene graphs to capture interactions among pedestrians, vehicles, and traffic elements across time. Social-STGCNN~\citep{mohamed2020socialstgcnnsocialspatiotemporalgraph} extends this idea by modeling dynamic social interactions, where nodes represent pedestrians and edges encode their evolving relationships, enabling accurate prediction in crowded environments.

Building on skeleton-based reasoning, ST-CrossingPose \citep{zhang2022st} employs spatial-temporal graph convolutions over pose sequences to jointly learn spatial structure and temporal evolution. Additional works incorporate contextual cues such as traffic signals and vehicle proximity~\citep{yang2022predicting}, or leverage graph-based visual reasoning frameworks~\citep{chen2021visual} to further enhance interaction-aware intent prediction. However, GNN-based methods often rely on predefined graph structures and may struggle to scale to highly dynamic and complex environments.

To overcome the limitations of both recurrent and graph-based models, Transformer-based approaches have recently emerged as a powerful framework for VRU intent prediction, owing to their ability to model long-range dependencies and complex interactions through self-attention mechanisms. CAPformer~\citep{loreto2021capformer} introduced a transformer encoder that jointly processes video frames and motion features, enabling global temporal reasoning without recurrence. Action-ViT~\citep{zhang2021actionvit} further demonstrated that Vision Transformers can effectively capture critical visual cues such as foot placement and gaze direction for action prediction. To improve reliability under uncertainty, TrEP~\citep{zhang2023trep} proposed an evidential Transformer framework that jointly predicts crossing probability and associated uncertainty, supporting risk-aware decision-making. Building on these advances, IntentFormer~\citep{sharma2024intentformer} introduced a multimodal transformer architecture that integrates visual, motion, and pose features, and jointly optimizes intent recognition, trajectory prediction, and uncertainty estimation through a composite learning objective. These methods highlight the advantage of Transformer-based models in providing a unified framework for modeling temporal dynamics, multimodal fusion, and agent interactions.

\subsubsection{Reasoning-Driven Behavior Understanding}
While early works primarily focus on intent recognition as a classification problem, recent advances in VLMs and LLMs increasingly emphasize reasoning capabilities, enabling models to interpret context, infer latent causes, and support proactive decision-making. For instance,  PedVLM~\citep{munir2025pedestrian} integrates a CLIP-based visual encoder with a T5 language model to fuse visual evidence with language-level reasoning, improving both prediction accuracy and interpretability. ClipCross~\citep{uziel2025optimizing} further demonstrates that embedding optimization is essential for adapting foundation models to subtle intention prediction tasks. Moving beyond specialized architectures, recent works investigate prompt-based reasoning with foundation models, where hierarchical prompts encoding contextual and temporal cues significantly enhance performance without additional training~\citep{azarmi2025pedestrian}. Open-source multimodal LLMs have also shown promising zero-shot generalization ability, as demonstrated by LLaMAPed~\citep{ham2024llamaped}, highlighting their potential for scalable deployment. Bridging visual perception and semantic reasoning, TCP~\citep{xiao2025tcp} introduces a text-guided cascade framework that incorporates textual priors to refine intention prediction across stages, effectively combining structured feature learning with high-level contextual knowledge. Together, these approaches represent a shift from purely data-driven prediction toward generative and reasoning-aware frameworks.

Beyond intent prediction, recent studies have extended reasoning capabilities toward more comprehensive scene understanding using VLMs and LLMs. For instance, \citet{jain2024semantic} leveraged LVLMs through visual question answering (VQA) to analyze complex traffic scenes, including infrastructure, vehicle interactions, environmental conditions, and VRU safety, demonstrating strong potential for multi-modal reasoning. Similarly, \citet{zhen2025multiviewphaseawarepedestrianvehicleincident} proposed a multi-view, phase-aware framework that decomposes pedestrian behavior into semantic stages and enables causal reasoning of pedestrian–vehicle interactions, reflecting a shift from intention prediction toward interpretable behavior analysis. Building on these capabilities, recent work further shows that reasoning can support practical traffic operations in closed-loop systems. For example, LimSim++ \citep{fu2024limsimclosedloopplatformdeploying} demonstrates that LLMs, when integrated with real-time visual and textual inputs, can not only understand dynamic scenes but also perform planning and control in simulation. This highlights the potential of reasoning-enabled agents to move beyond passive understanding toward actionable decision-making for proactive VRU protection.

In summary, intent recognition has evolved from sequential modeling to interaction-aware and, more recently, reasoning-driven paradigms. Early approaches focus on temporal pattern learning, while later methods incorporate social interactions and global context. Recent advances in vision-language frameworks further enable semantic reasoning and uncertainty modeling, marking a transition from purely predictive systems to more interpretable and proactive behavior understanding frameworks.

\begin{table*}[!ht]
\renewcommand{\arraystretch}{1.3}
\begin{footnotesize}
\begin{center}
\caption{Representative Methods for VRU Intent Recognition and Reasoning (T: Trajectory; P: Pose; V: Visual Content)}
\label{tab:vru_intent_methods}
\begin{tabularx}{0.98\textwidth}{@{}llllX@{}}
\toprule
\textbf{Category} & \textbf{Method} & \textbf{Year} & \textbf{Input} & \textbf{Key Features and Contributions} \\

\midrule

\multirow{6}{*}{\shortstack[l]{RNN-based\\Models}} 
& VRUNet~\citep{ranga2020vrunet} &2020& P, V & Multi-task LSTM jointly predicts pedestrian action, crossing intent, and future trajectory. \\
& Coupled Intent-Action~\citep{yao2021coupling} &2021& T, V & Recurrent encoding of present and predicted actions to refine crossing intent. \\
& PIP-Net~\citep{azarmi2024pip} &2024& T, P, V & Recurrent-attentive enables robust forecasting several seconds ahead under occlusion. \\
& PCPA~\citep{kotseruba2021benchmark} &2021& T, P, V & Modality-specific temporal attention improves crossing prediction and interpretability. \\
& MCIP~\citep{ham2022mcip} &2022& T, P, V & Attention-based multimodal fusion enhances robustness to missing inputs. \\
& CIPF~\citep{ham2023cipf} &2023& T, P, V & Temporal encoding and fusion across modalities improve crossing prediction accuracy. \\

\midrule

\multirow{7}{*}{\shortstack[l]{GNN-based\\Models}} 
& Pedestrian Graph~\citep{cadena2019pedestrian} &2019& P, V & Early GCN predicting crossing intent from pedestrian skeleton graphs. \\
& Pedestrian Graph+~\citep{cadena2022pedestrian} &2022& P, V & Faster, optimized graph construction for real-time pedestrian crossing prediction. \\
& SpatioTemporal GCN~\citep{liu2020spatiotemporal} &2020& T, V & Models evolving scene relationships for crossing intent and social reasoning. \\
& Social-STGCNN~\citep{mohamed2020socialstgcnnsocialspatiotemporalgraph} &2020& T, V & Dynamic social graph learning for crossing and trajectory prediction. \\
& ST-CrossingPose~\citep{zhang2022st} &2022& T, P & Spatio-temporal skeleton learning for real-time crossing behavior forecasting. \\
& Yang et al.~\citep{yang2022predicting} &2022& T, P, V & Combines pose and scene context for enhanced crossing intent prediction. \\
& Chen et al.~\citep{chen2021visual} &2021& T, P, V & Visual relational reasoning for intent prediction from ego-view. \\

\midrule

\multirow{4}{*}{\shortstack[l]{ViT-based\\Models}} 
& CAPformer~\citep{loreto2021capformer} &2021& T, P, V & Captures global temporal behavior patterns with attention-based modeling. \\
& Action-ViT~\citep{zhang2021actionvit} &2021& T, P, V & Highlights critical visual cues for pedestrian action prediction. \\
& TrEP~\citep{zhang2023trep} &2023& T, V & Jointly predicts crossing probability and uncertainty for risk-aware decision-making. \\
& IntentFormer~\citep{sharma2024intentformer} &2024& T, P, V & Unified multimodal transformer for joint intent, trajectory, and uncertainty estimation. \\

\midrule

\multirow{5}{*}{\shortstack[l]{LLM-based\\Models}}
& PedVLM~\citep{munir2025pedestrian} &2025& T, V & Vision-language fusion using CLIP and T5 for semantic-aware intent prediction. \\
& ClipCross~\citep{uziel2025optimizing} &2025& V & Optimized visual-language embeddings for fine-grained pedestrian intention prediction. \\
& \citet{azarmi2025pedestrian} &2025& V & Hierarchical prompt design enables context-aware reasoning without additional training. \\
& LLaMAPed~\citep{ham2024llamaped} &2024& V & Multimodal LLM enabling zero-shot intent prediction with strong generalization. \\
& TCP~\citep{xiao2025tcp} &2025& T, V & Text-guided cascade framework using semantic priors for multi-stage intent refinement. \\

\bottomrule
\end{tabularx}
\end{center}
\end{footnotesize}
\end{table*}

\section{Open Challenges and Future Directions}
\label{sec:challenges}

\begin{figure*}[!ht]
	\centering
	\includegraphics[width=.95\textwidth]{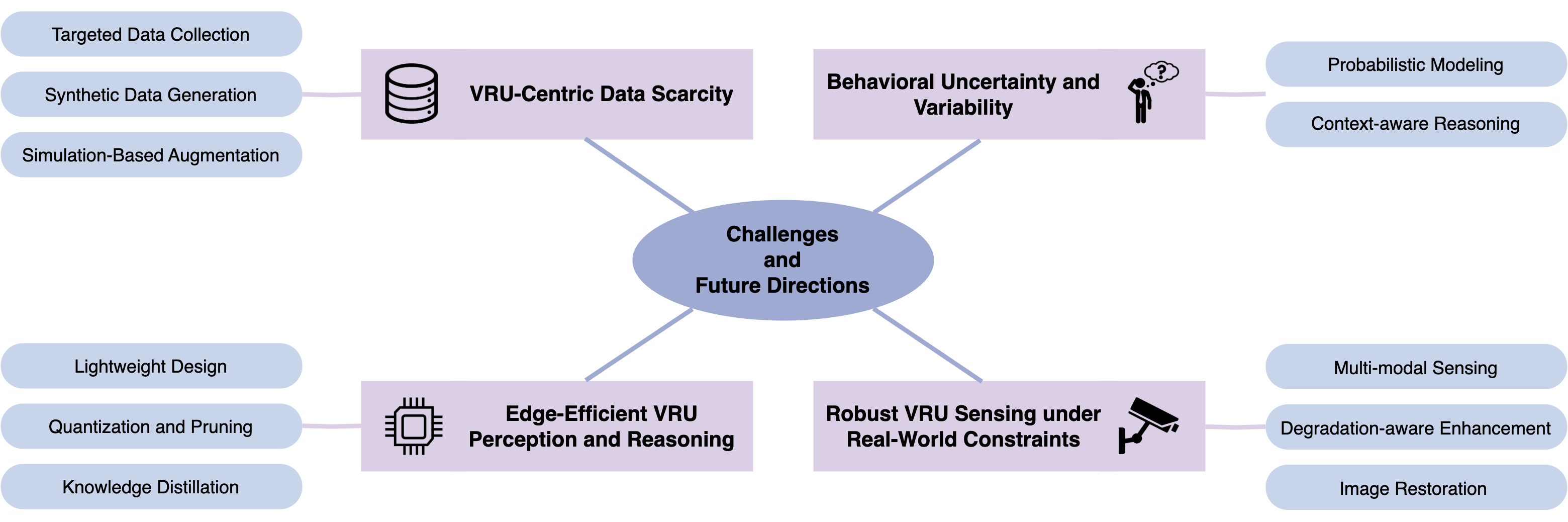}
	\caption{Open challenges and future research directions for camera-based VRU sensing and reasoning, organized into four categories. Each category summarizes key problems and corresponding emerging solutions discussed in Section~\ref{sec:challenges}.}
	\label{fig:challenges}
\end{figure*}

As intelligent systems for VRU safety continue to advance, several key challenges still hinder their reliable deployment in real-world environments. Despite progress in detection, motion modeling, and behavior understanding, achieving robust performance under diverse conditions remains difficult. In particular, current systems are limited by four factors: data scarcity and imbalance, the inherent uncertainty of VRU behaviors, the trade-off between efficiency and accuracy for real-time deployment, and sensitivity to environmental variability and sensor degradation. These challenges affect the entire pipeline from perception to reasoning and constrain the realization of proactive VRU safety. In this section, as illustrated in Figure~\ref{fig:challenges}, we summarize these challenges and discuss potential research directions toward more reliable and scalable VRU safety systems.

\subsection{VRU-Centric Data Scarcity}
A fundamental challenge in AI-enabled VRU safety is the scarcity of data that adequately supports the full pipeline of detection, prediction, and reasoning. This limitation primarily arises from two aspects: the underrepresentation of diverse VRU types and the lack of safety-critical interaction scenarios.

First, existing datasets are often imbalanced across VRU categories. While common classes such as pedestrians and cyclists are relatively well represented, other VRUs, including children, elderly individuals, and micro-mobility users, remain underrepresented \citep{katare2024analyzingmitigatingbiasvulnerable}. This imbalance impacts all three tasks: detection models may fail to recognize less frequent VRU types, trajectory prediction models may not capture their distinct motion patterns, and reasoning models may misinterpret their behaviors. As a result, system reliability degrades in heterogeneous and real-world urban environments.

Second, current datasets lack sufficient coverage of behaviorally rich and safety-critical VRU scenarios, such as sudden crossings, hesitation, occlusion-heavy interactions, and complex multi-agent negotiations. These long-tail events are crucial for proactive safety, as they require models not only to detect VRUs but also to anticipate their future trajectories and infer potential risks. However, such scenarios are inherently rare and difficult to capture at scale, limiting model robustness in high-risk conditions.

To address these challenges, recent efforts have explored both targeted data collection and advanced augmentation techniques. For example, the Intersection Safety Challenge \citep{usdotsafetychallenge} employs controlled setups using mannequins mounted on mobile platforms to simulate diverse VRU behaviors, enabling the collection of underrepresented cases. In addition, recent datasets have begun to emphasize richer behavioral and contextual annotations. The WTS dataset \citep{kong2024wtspedestriancentrictrafficvideo}, for instance, introduces pedestrian-centric traffic videos with dense textual descriptions and multimodal annotations, supporting fine-grained analysis of VRU behavior and interaction context. Such efforts move beyond traditional detection-focused datasets toward enabling joint perception, prediction, and reasoning. In parallel, synthetic data generation and simulation-based augmentation provide scalable solutions to enrich both VRU diversity and safety-critical scenarios. Scenario Diffusion \citep{pronovost2023scenario}, for instance, introduces a diffusion-based framework capable of generating controllable, safety-critical driving scenarios, demonstrating the potential of generative models to effectively complement real-world datasets.

\subsection{Behavioral Uncertainty and Variability}

A critical challenge in AI-enabled VRU safety is the inherent uncertainty and variability of VRU behaviors, which makes reliable prediction and reasoning difficult. Unlike vehicles that typically follow structured rules, VRUs exhibit diverse, spontaneous, and often ambiguous behaviors influenced by individual intent, social interactions, and environmental context. From a prediction perspective, VRU motion is inherently multi-modal, where similar observations can lead to multiple plausible future trajectories. From a reasoning perspective, intent inference is equally challenging, as identical visual cues (e.g., standing near a curb) may correspond to different underlying behaviors depending on context. Such ambiguity limits the ability of current systems to support proactive safety, which requires early and reliable anticipation of potential risks.

To address these challenges, recent research has increasingly explored probabilistic and context-aware modeling approaches. In particular, generative models, such as diffusion and flow-based methods, provide a natural framework for capturing the multi-modal nature of VRU behaviors by modeling distributions over possible future trajectories and actions. These approaches enable systems to reason over multiple hypotheses rather than relying on single deterministic predictions, thereby improving robustness under uncertainty. In parallel, VLMs and LLMs introduce high-level semantic reasoning capabilities by incorporating contextual and relational information, further enhancing interpretability and supporting more informed intent inference. Together, these advances represent a shift toward uncertainty-aware and reasoning-driven frameworks for more reliable VRU safety systems.

\subsection{Edge-Efficient VRU Perception and Reasoning}

Deploying AI-based VRU safety systems in real-time edge environments presents a significant challenge, as models must balance computational efficiency with the preservation of critical perception and behavior cues. Unlike generic vision tasks, VRU safety relies on accurately capturing fine-grained spatial and temporal information, making aggressive model compression prone to degrading performance in safety-critical scenarios. This challenge affects the entire pipeline. For detection, lightweight models must remain robust to small-scale and occluded VRUs. For trajectory prediction, efficiency constraints may limit the modeling of temporal dependencies and interaction dynamics. For behavior understanding, resource limitations restrict the deployment of models capable of contextual and semantic reasoning. These trade-offs directly hinder proactive safety, where timely and reliable inference is essential.

To address these challenges, recent research has shifted from generic compression toward task-aware and semantics-preserving optimization. Lightweight architectures reduce computational cost through efficient design strategies such as depthwise separable convolutions and compound scaling \citep{Setyanto2024}, but must be carefully designed to retain features critical for small-object detection and motion modeling. In addition, model compression techniques, including pruning, quantization, and knowledge distillation, play a central role in enabling efficient deployment. Structured pruning removes redundant parameters while maintaining model effectiveness, quantization reduces memory footprint and inference latency, and knowledge distillation allows compact models to inherit key representations from larger networks. Recent studies show that such optimized models can achieve competitive accuracy with significantly reduced inference time \citep{cai2021yolobile}.

\subsection{Robust VRU Sensing under Real-World Constraints}

The reliability of AI-enabled VRU safety systems is fundamentally constrained by real-world sensing conditions, where environmental variability and sensor degradation can significantly impair perception and downstream reasoning. Unlike controlled settings, real-world deployments must operate under diverse lighting, weather, and infrastructure conditions, which directly affect the visibility and detectability of VRUs. These challenges are particularly critical for VRUs due to their small physical scale, deformable shapes, and frequent occlusions. Under adverse conditions such as low illumination, rain, fog, and glare, visual sensors, especially RGB cameras, often suffer from reduced contrast, motion blur, and partial visibility, leading to degraded detection performance \citep{vargas2021overview}. Such degradation not only affects detection accuracy but also propagates through the pipeline.

To improve robustness, multi-modal sensing strategies have been widely explored to complement visual information under challenging conditions. While effective, these approaches introduce additional complexities, including sensor calibration, synchronization, and increased computational cost, which can hinder real-time deployment. Alternatively, recent research has demonstrated that algorithmic approaches can enhance robustness using vision-based sensing alone. For example, the Edge-MuSE system \citep{liu2023real} performs multi-task environmental perception directly from camera inputs, such as visibility estimation, dehazing, road segmentation, and surface condition classification, allowing the system to compensate for degraded visual inputs and improve perception reliability under adverse conditions.

Beyond short-term environmental variability, long-term sensor degradation further affects system performance. In outdoor deployments, sensors are subject to wear, contamination, and misalignment, leading to gradual degradation in data quality. To address these issues, recent work has explored restoration and self-adaptive approaches. Universal image restoration models, such as AirNet \citep{li2022all}, and diffusion-based methods like DiffUIR \citep{zheng2024selective}, aim to recover degraded visual inputs without prior knowledge of corruption types. These approaches offer a promising direction for maintaining reliable perception in dynamic and long-term real-world environments.

\section{Conclusion}
\label{sec:conclusion}
In this survey, we presented a comprehensive review of recent advances in vision-based tasks for improving VRU safety.  Moving beyond traditional detection-focused studies, we organized existing work into a unified pipeline consisting of visual perception (detection and classification), motion modeling (tracking and trajectory prediction), and behavior understanding (intent recognition and reasoning). This structured perspective highlights how these interconnected components collectively enable proactive and context-aware VRU protection in intelligent transportation systems. We further discussed how these proactive protection methods can be integrated into existing transportation systems to enhance VRU safety through the cooperation of VRU-centric infrastructure. Our review highlights the growing role of advanced AI paradigms, such as ViTs, LLMs, and Diffusion Models, in enhancing representation learning, uncertainty modeling, and semantic reasoning. These emerging methods offer promising solutions for modeling complex behaviors and interactions in dynamic traffic environments, yet they remain underexplored in existing literature. 

Despite significant progress, several key challenges persist, including data scarcity and imbalance, behavioral uncertainty, efficiency constraints for real-time deployment, and sensitivity to real-world sensing variability. Addressing these challenges is essential for developing reliable, scalable, and robust VRU safety systems. Future research should focus on advancing data-efficient, uncertainty-aware, and computationally efficient models, while improving generalization across diverse environments.

Overall, by bridging visual sensing, motion modeling, and reasoning with practical deployment considerations, next-generation AI systems have the potential to enable more proactive, adaptive, and inclusive VRU safety in real-world transportation systems.

\section*{Acknowledgment}
This work was supported by the Pacific Northwest Transportation Consortium (PacTrans), University of Washington.

{\small
\bibliographystyle{IEEEtranN}
\bibliography{IEEEabrv,reference}
}

\newpage

\section{Biography Section}
\begin{IEEEbiography}[{\includegraphics[width=1in,height=1.25in,clip,keepaspectratio]{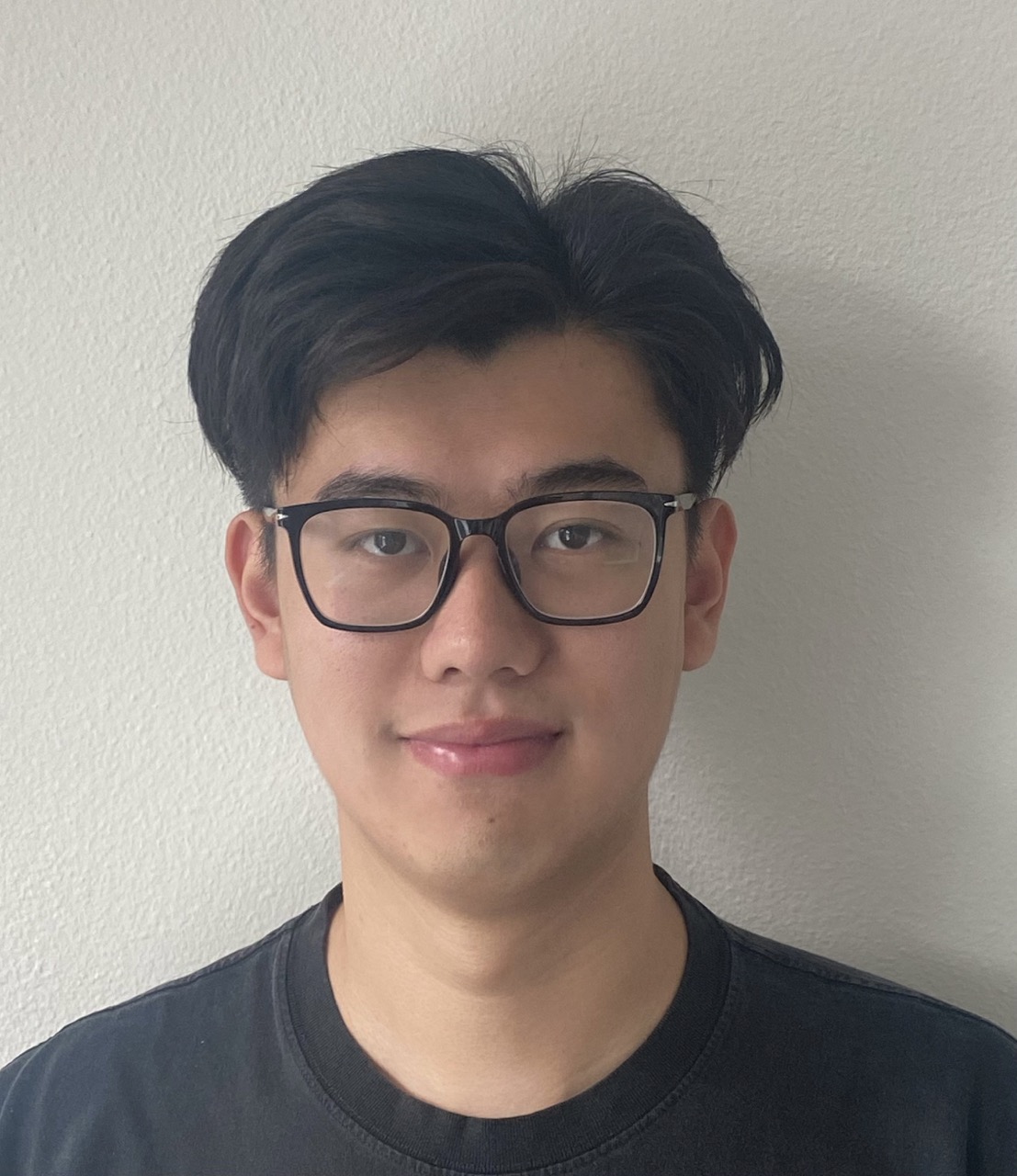}}]{Shucheng Zhang} is currently pursuing the Ph.D. degree in the Smart Transportation Application and Research (STAR) Lab, Department of Civil and Environmental Engineering at the University of Washington. Prior to joining the STAR Lab, Shucheng earned his M.S. in Mechanical Engineering from Duke University. His research interests include computer vision, intelligent transportation systems, and autonomous vehicles, with a focus on developing innovative solutions to enhance road safety and transportation automation.
\end{IEEEbiography}

\vspace{-1cm} 

\begin{IEEEbiography}[{\includegraphics[width=1in,height=1.25in,clip,keepaspectratio]{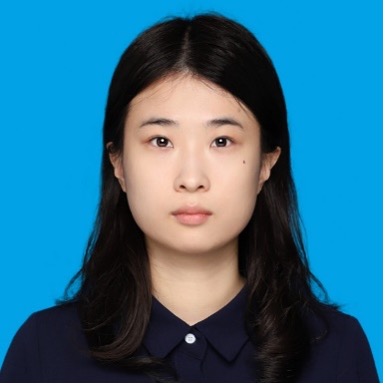}}]{Yan Shi} is a Ph.D. student in the Department of Civil Engineering at the University of Washington. Her research focuses on intelligent transportation systems, vision-language models for traffic safety, and reasoning approaches such as chain-of-thought. She has served as a reviewer for journals and conferences including IEEE Transactions on Intelligent Transportation Systems (ITS), IEEE Transactions on Image Processing (TIP), Transportation Policy, ACM Multimedia (ACM MM), and the Transportation Research Board (TRB).
\end{IEEEbiography}

\vspace{-1cm} 

\begin{IEEEbiography}[{\includegraphics[width=1in,height=1.25in,clip,keepaspectratio]{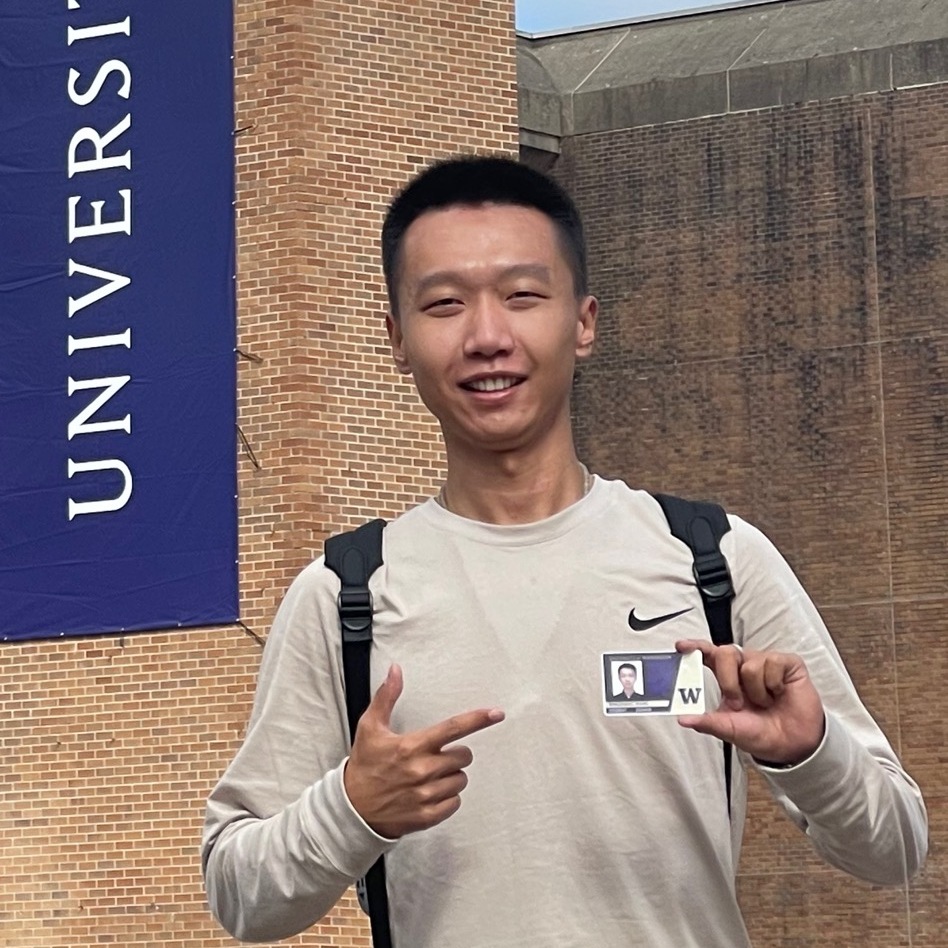}}]{Bingzhang Wang} received double B.S. degrees in Mechanical Engineering from Shanghai Jiao Tong University (2020) and Software Engineering from Peking University (2022), and a M.S. degree in Transportation Engineering from the University of Washington (2024). He is currently a Ph.D. student at the Smart Transportation Research and Application Lab (STAR Lab), Department of Civil and Environmental Engineering, University of Washington. He served as a reviewer for IEEE Transactions on Intelligent Transportation Systems, IEEE International Conference on Intelligent Transportation Systems, and other journals and conferences. His research interests lie in representation learning, deep generative models, and multimodal data analytics for intelligent and autonomous transportation systems.
\end{IEEEbiography}

\vspace{-1cm} 

\begin{IEEEbiography}[{\includegraphics[width=1in,height=1.25in,clip,keepaspectratio]{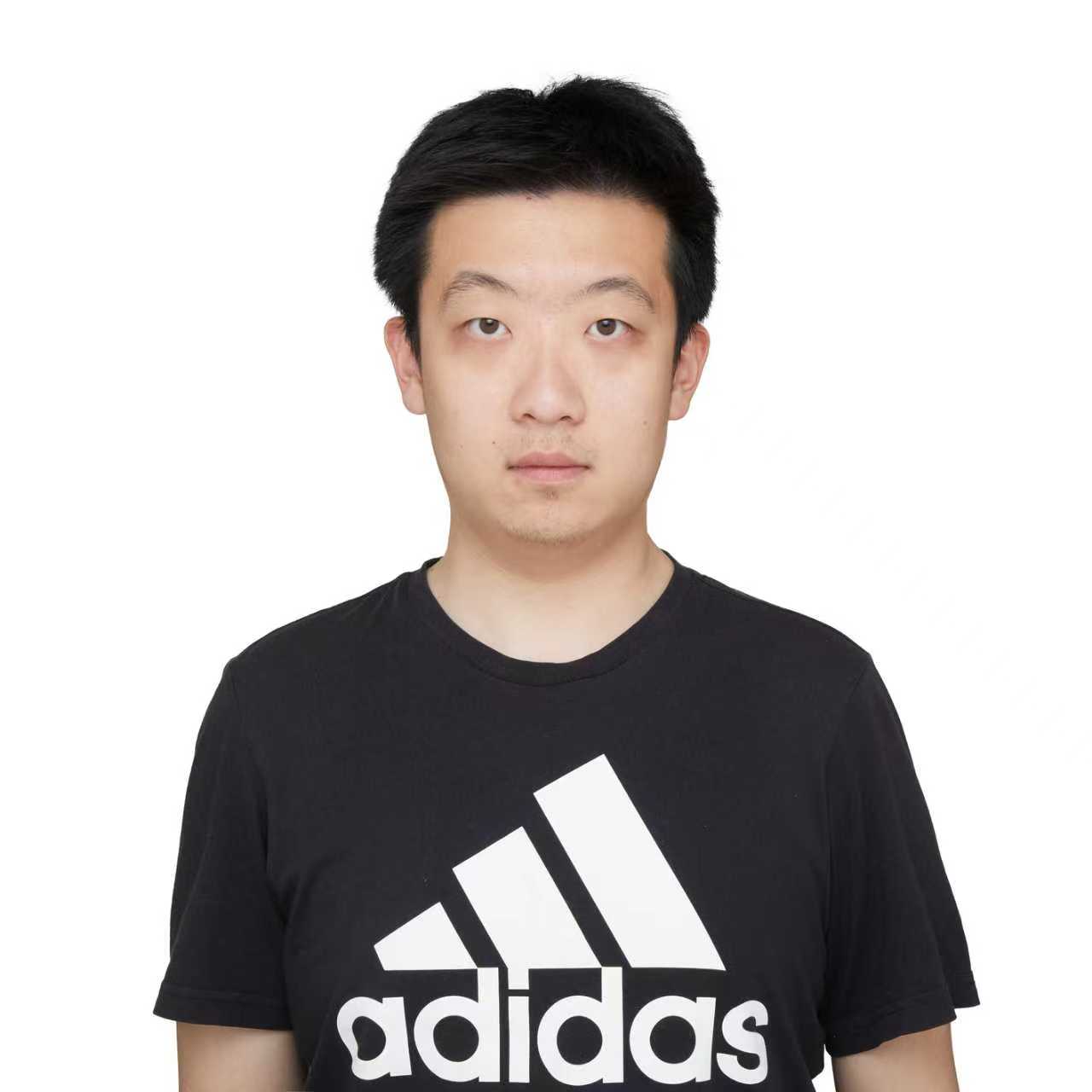}}]{Yuang Zhang} received the bachelor's and master's degrees from the Department of Automation, Tsinghua University. He is currently working toward the Ph.D. degree with the Department of Civil and Environmental Engineering, University of Washington, Seattle, Washington. He has published papers in IEEE TRANSACTIONS ON MOBILE COMPUTING (TMC), IEEE ICRA and IEEE IST, etc. His research interests are focused on generative AI, computer vision and autonomous driving.
\end{IEEEbiography}

\vspace{-1cm} 

\begin{IEEEbiography}[{\includegraphics[width=1in,height=1.25in,clip,keepaspectratio]{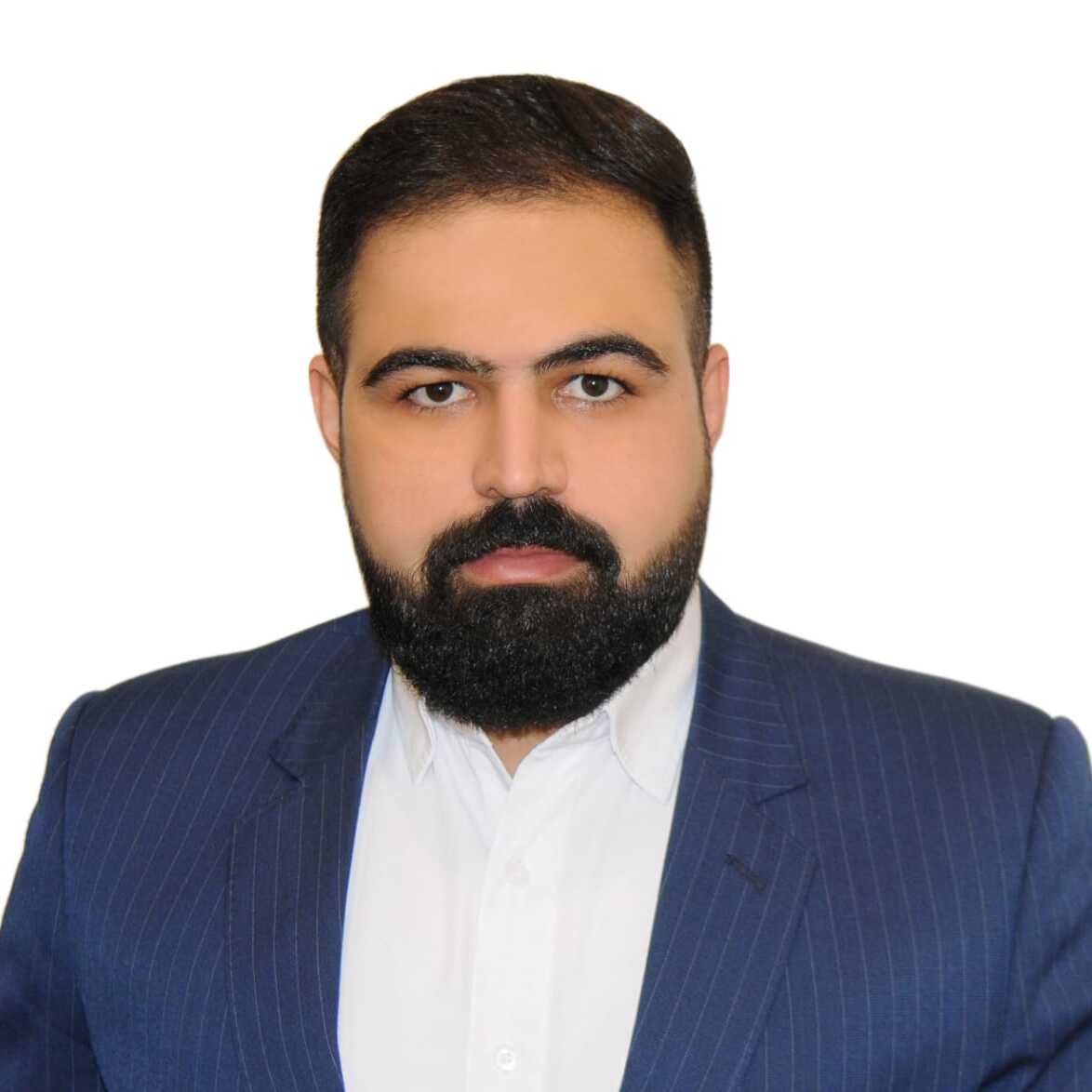}}]{Mehrdad Nasri} is a Graduate Research Assistant and Ph.D. candidate in Transportation Engineering at the University of Washington, focusing on transportation safety. He earned an M.Sc. in Transportation Engineering from the University of Tehran. His research leverages AI, machine learning, deep learning, and advanced statistical methods to extract safety insights from emerging data sources including connected-vehicle data, street view images, and multimodal sensor networks. Committed to bringing research into practice, Mehrdad develops tools and guidelines that reduce crash risk for all road users, from transit riders to pedestrians and cyclists. He has authored several peer-reviewed papers on data-driven safety improvements.
\end{IEEEbiography}

\begin{IEEEbiography}[{\includegraphics[width=1in,height=1.25in,clip,keepaspectratio]{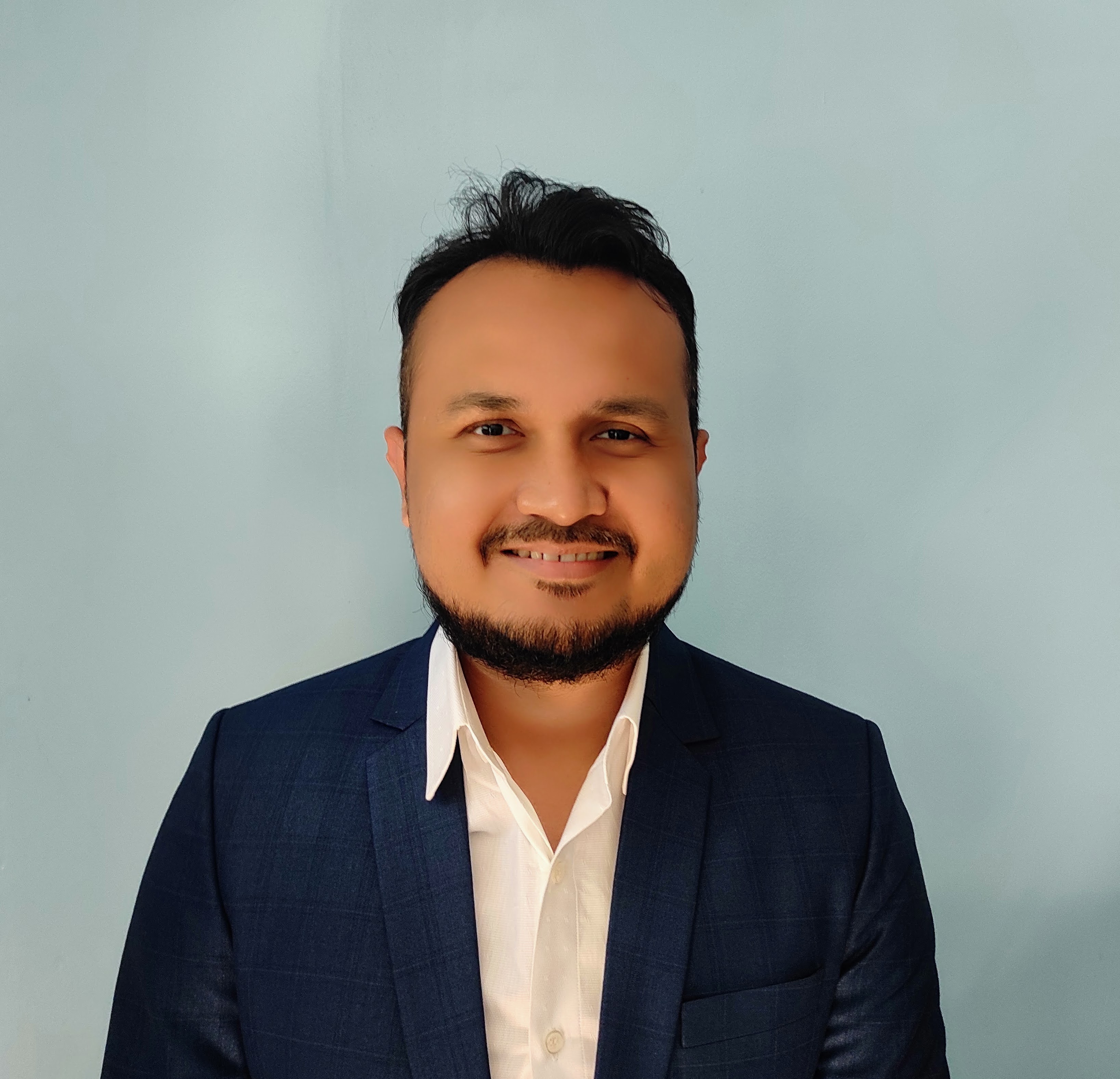}}]{Muhammad Monjurul Karim} is a Postdoctoral Scholar at the Smart Transportation Applications and Research Lab (STAR Lab) at the University of Washington in Seattle. He earned his Ph.D. in Civil Engineering from Stony Brook University and an M.S. in Systems Engineering from the Missouri University of Science and Technology.
He is interested in solving large-scale visual recognition and prediction problems by developing novel deep learning approaches. His research experience includes, but is not limited to traffic accident anticipation, risk localization, object detection/segmentation, tracking, visual attention, and remote sensing.
\end{IEEEbiography}

\begin{IEEEbiography}[{\includegraphics[width=1in,height=1.35in,clip,keepaspectratio]{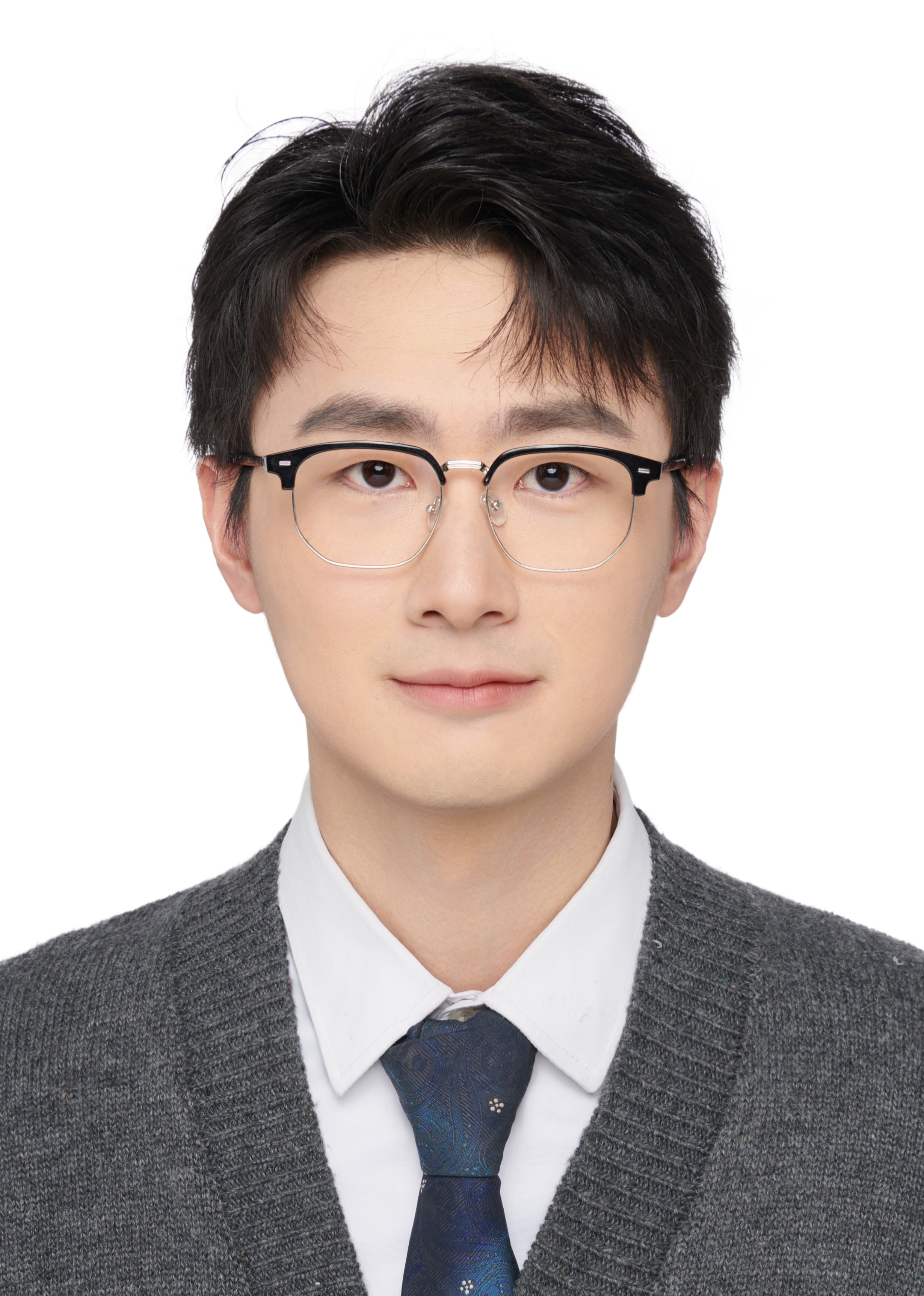}}]{Kehua Chen} received a B.S. degree in Civil Engineering from Chongqing University and a dual M.S. degree in Environmental Sciences from the University of Chinese Academy of Sciences and the University of Copenhagen. He earned his Ph.D. in Intelligent Transportation from the Hong Kong University of Science and Technology in 2024. Currently, he is a postdoctoral scholar at the Smart Transportation Applications and Research (STAR) Lab at the University of Washington. His research interests encompass urban and sustainable computing, as well as autonomous driving.
\end{IEEEbiography}

\begin{IEEEbiography}[{\includegraphics[width=1in,height=1.25in,clip,keepaspectratio]{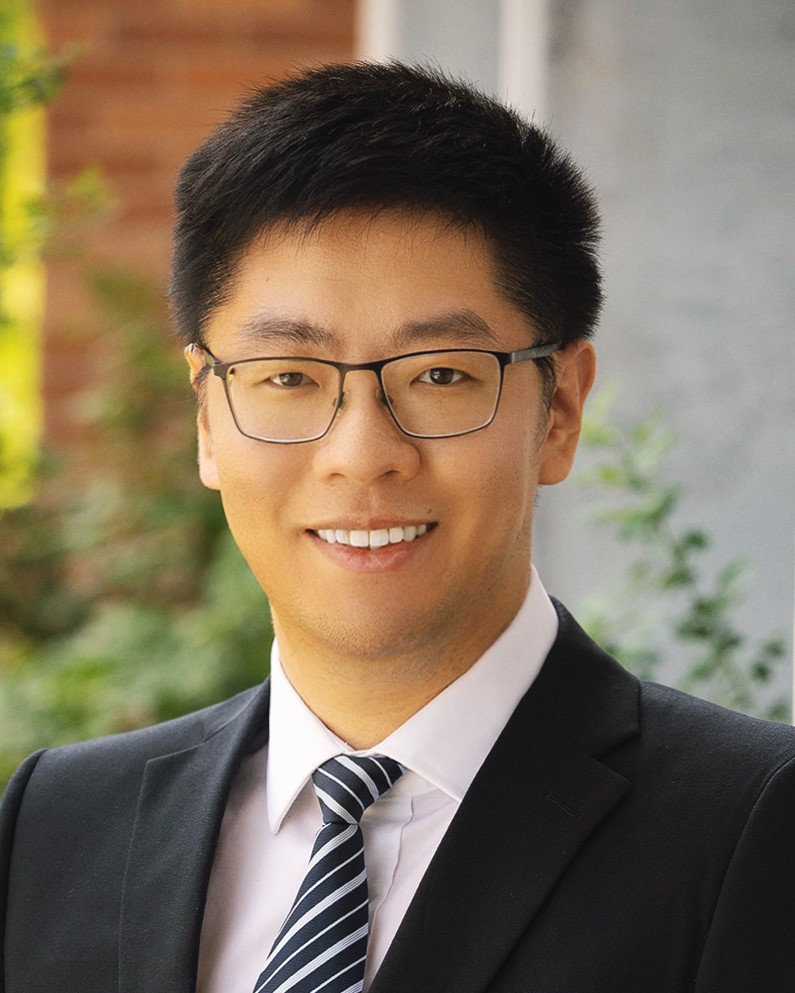}}]{Chenxi Liu} is an Assistant Professor in Civil and Environmental Engineering at the University of Utah. He received his B.S. from Tsinghua University (2017), and M.S. (2020) and Ph.D. (2024) from the University of Washington. His research develops situation-aware, customized machine intelligence for safe and resilient transportation systems. He integrates advanced traffic sensing (environment, 3D object, crowds, cooperative) with edge computing for various transportation applications. He also explores generative AI, including LLM-based solutions, to enhance V2X communications.
\end{IEEEbiography}

\begin{IEEEbiography}[{\includegraphics[width=1in,height=1.25in,clip,keepaspectratio]{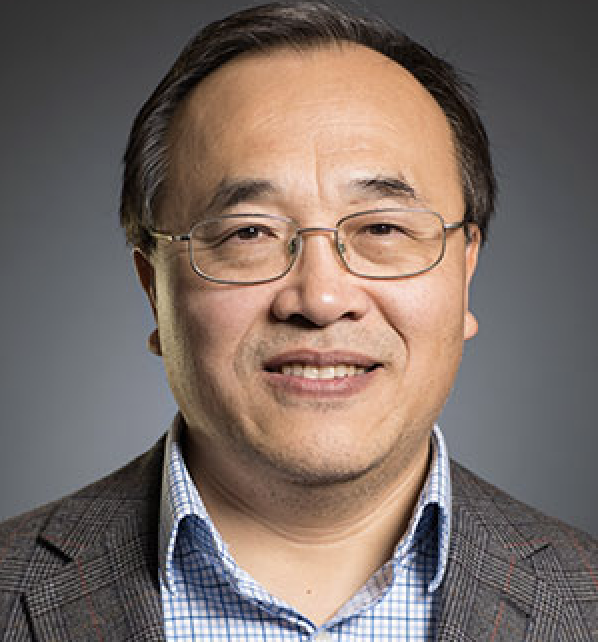}}]{Yinhai Wang} received the master’s degree in computer science from the University of Washington (UW) and the Ph.D. degree in transportation engineering from The University of Tokyo in 1998. He is currently a Professor in transportation engineering and the Founding Director of the Smart Transportation Applications and Research Laboratory (STAR Lab), UW. He also serves as the Director of the Pacific Northwest Transportation Consortium (PacTrans), U.S. Department of Transportation, University Transportation Center for Federal Region 10.
\end{IEEEbiography}

\vfill

\end{document}